\documentclass[runningheads]{llncs}
\usepackage[T1]{fontenc}
\usepackage{graphicx}
\usepackage{amsmath}
\usepackage{amsfonts}
\usepackage[para,online,flushleft]{threeparttable}
\usepackage{float}            
\usepackage{subfig}
\usepackage{overpic}
\usepackage{booktabs}      
\usepackage{multirow}     
\usepackage{siunitx}      
\usepackage{xcolor}      
\usepackage{caption}

\definecolor{myblue}{RGB}{0, 0, 200}      
\definecolor{mypurple}{RGB}{128, 0, 128}  
\definecolor{mygreen}{RGB}{0, 128, 0}    
\definecolor{myred}{RGB}{200, 0, 0}      

\newcommand{\diffcolor}[1]{%
  \ifdim #1pt < 0pt \textcolor{myblue}{#1}%
  \else\ifdim #1pt > 0pt \textcolor{mypurple}{#1}%
  \else #1%
  \fi\fi
}

\newcommand{\bestval}[1]{\textcolor{mygreen}{\textbf{#1}}}
\newcommand{\worstval}[1]{\textcolor{myred}{#1}}

\begin{document}
\title{A Benchmark Study of Deep Reinforcement Learning Algorithms for the Container Stowage Planning Problem}
\titlerunning{Benchmarking DRL for Container Stowage Planning}
\author{Yunqi Huang\inst{1,2} \and
Nishith Chennakeshava\inst{2} \and
Alexis Carras\inst{2} \and
Vladislav Neverov\inst{2} \and
Wei Liu\inst{1} \and
Aske Plaat\inst{1} \and
Yingjie Fan\inst{1}}
\authorrunning{Huang Y. et al.}
%
\institute{Leiden University, Leiden Institute of Advanced Computer Science, \\ Leiden, The Netherlands \\ \email{\{w.liu,a.plaat,y.fan\}@liacs.leidenuniv.nl} \and
loadmaster.ai, Rotterdam, The Netherlands
\email{\{yunqi,nish,alexis,vladislav\}@loadmaster.ai}
}

\maketitle              
\begin{abstract}

Container stowage planning (CSPP) is a critical component of maritime transportation and terminal operations, directly affecting supply chain efficiency. Owing to its complexity, CSPP has traditionally relied on human expertise. While reinforcement learning (RL) has recently been applied to CSPP, systematic benchmark comparisons across different algorithms remain limited. To address this gap, we develop a Gym environment that captures the fundamental features of CSPP and extend it to include crane scheduling in both multi-agent and single-agent formulations. Within this framework, we evaluate five RL algorithms: DQN, QR-DQN, A2C, PPO, and TRPO under multiple scenarios of varying complexity. The results reveal distinct performance gaps with increasing complexity, underscoring the importance of algorithm choice and problem formulation for CSPP. Overall, this paper benchmarks multiple RL methods for CSPP while providing a reusable Gym environment with crane scheduling, thus offering a foundation for future research and practical deployment in maritime logistics.
\keywords{Container Stowage Planning \and Deep Reinforcement Learning \and Multi-Agent Systems \and Crane Scheduling}

\end{abstract}
\section{Introduction}
Container ports are vital hubs in the global supply chain, and the efficiency of container stowage planning directly impacts terminal operations and turnaround times~\cite{HE2023101864,zhou2022emerging}. The Container Stowage Planning Problem (CSPP), which seeks an optimal loading sequence under numerous physical and logistical constraints, is NP-hard and has traditionally relied on the expertise of human planners \cite{avriel1993exact,jensen2018container}. However, manual planning has become a bottleneck as vessel sizes and throughput demands increase, highlighting the need for automation, while ports adopting AI and automation have reported faster cargo turnover \cite{shen2017deep,:/content/books/9789211065923}. Recent studies have applied reinforcement learning (RL) to CSPP, but often focus on single algorithms, lacking comprehensive benchmarks across different algorithms \cite{jiang2021new,wei2021optimization}.

An aspect of stowage planning is the joint optimization with equipment scheduling, particularly for multi-crane operations, as siloed optimization can yield limited overall improvement \cite{hsu2021joint}. Research into jointly solving CSPP and crane scheduling is relatively scarce \cite{kizilay2021comprehensive}, especially using RL methods. This joint problem can be formulated as either a single-agent problem, where one agent selects both containers and cranes, or a multi-agent problem where each agent representing a crane makes container selection decisions respectively. The impact of these two formulations on the performance of different RL algorithms remains to be studied.

To address these gaps, we develop the Stowage Planning Gym Environment (SPGE), a customizable RL platform that captures the core challenges of CSPP and can be extended to incorporate crane scheduling. SPGE abstracts vessels and yards as 3D grids of slots, each characterized by position, occupancy, and group attributes, enabling constraints from shipping rules to be enforced. This slot-based abstraction allows flexible scaling of problem complexity, reproducibility, and compatibility with standard RL libraries. We further extend SPGE to incorporate crane scheduling through a joint container–crane model (SPGE-MC) and a multi-agent variant (SPAEC).
Using this platform, we conduct a comprehensive benchmark of five RL algorithms (DQN, QR-DQN, A2C, PPO, and TRPO) across a series of scenarios with increasing complexity. Our experiments highlight distinct performance gaps as complexity increases, underscoring the strong dependence of algorithm effectiveness on scenario characteristics in CSPP. These findings also validate our Gym environment as a reusable foundation for future research.

The remainder of this paper is structured as follows. 
Section 2 introduces the Container Ship Stowage Planning Problem. 
Section 3 reviews related work. 
Section 4 outlines the preliminaries. 
Section 5 presents the methodology and the experimental results. 
Finally, Section 6 concludes the paper and provides future directions.

\section{Container Ship Stowage Planning Problem}

\begin{figure}[htbp]
    \centering
    \includegraphics[width=1\linewidth]{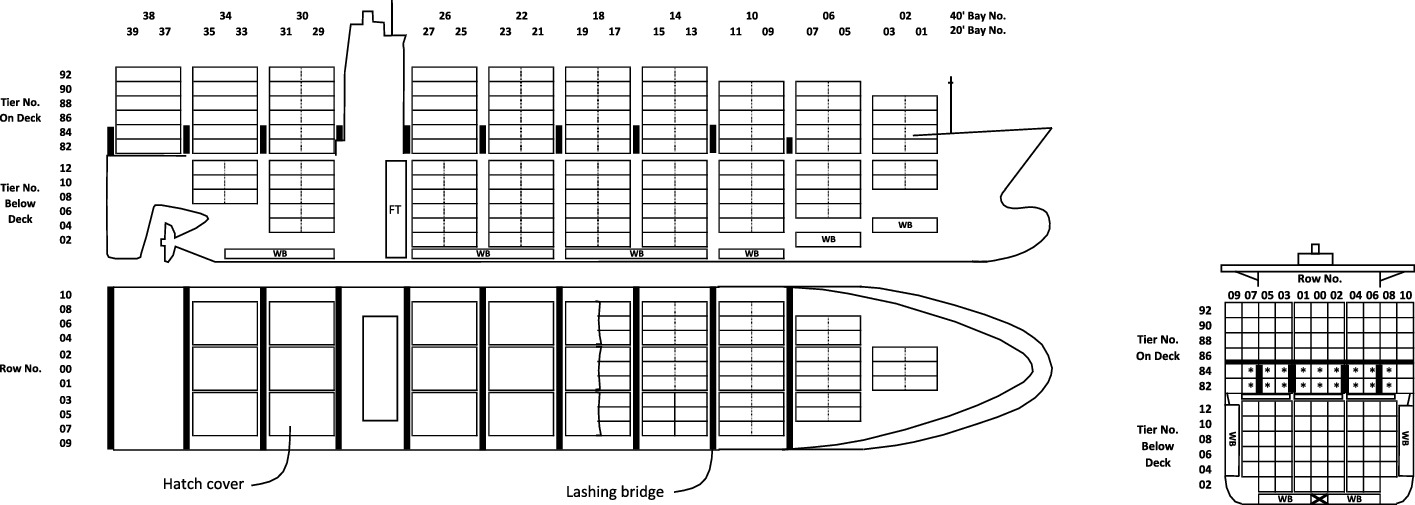}
    \caption{\textit{Vessel structure}\cite{VANTWILLER2024841}.}
    \label{fig:vessel_structure}
\end{figure}

\begin{figure}[htbp]
    \centering
    \includegraphics[width=0.8\linewidth]{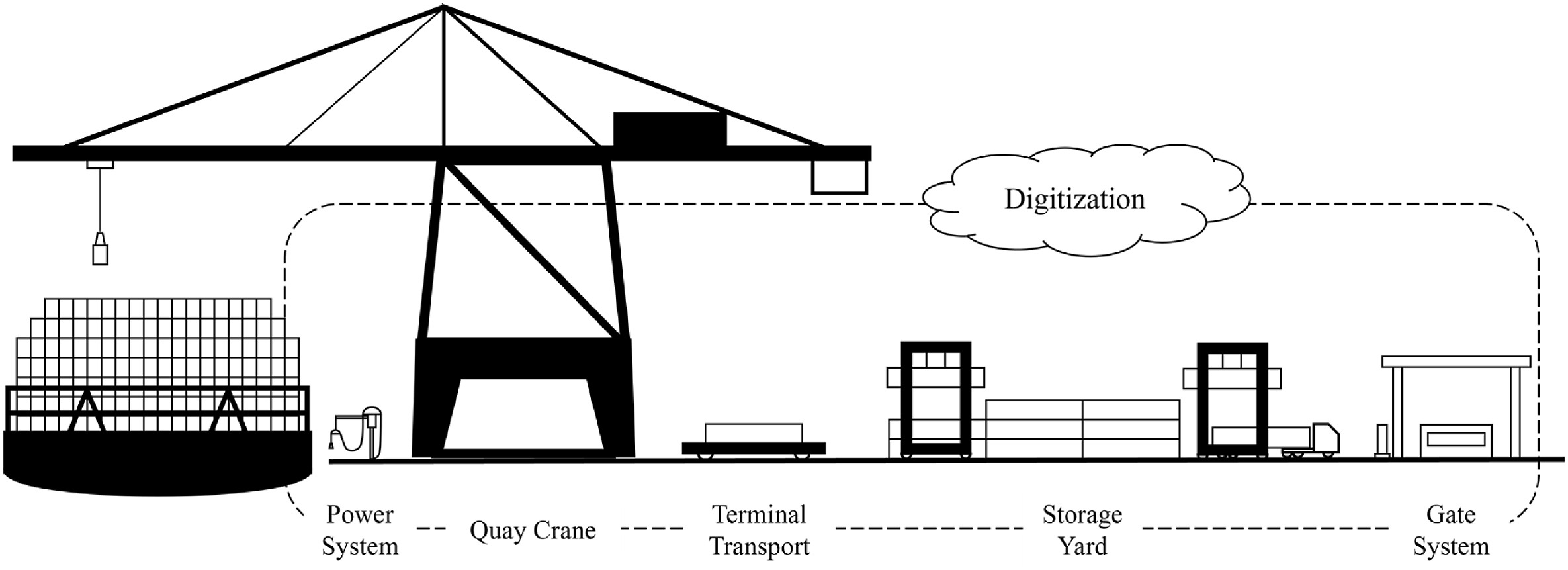}
    \caption{\textit{Stowage operation process}\cite{zhou2022emerging}.}
    \label{fig:operation precess}
\end{figure}

The fundamental unit for container stowage is the slot, defined by a (bay, row, tier) coordinate system on both the vessel and in the yard, as shown in Figure \ref{fig:vessel_structure}. The stowage operation, depicted in Figure \ref{fig:operation precess}, involves moving containers from the storage yard to the vessel using handling equipment and quay cranes. A key challenge in this process is the "shifter" or reshuffling operation, which occurs when a target container is blocked by others in the yard stack, requiring additional movements. The overall process for a single container includes potential shifters, transport to the quay, and loading by the crane onto the vessel. In modern terminals, this can involve multiple cranes operating in parallel.

Our primary optimization objectives are minimizing shifters and total operation duration. Furthermore, we extend the problem to include crane scheduling, where for $k$ cranes in set $CR$, the goal is to find an optimal sequence of pairs $\langle (p_1, o_1), \dots, (p_m, o_m) \rangle$, where $p_i$ is the container and $o_i \in CR$ is the assigned crane.

The CSPP involves placing $m$ containers from a set $C$ into $n$ vessel slots $S$ \cite{AMBROSINO200481}. The problem can be framed as a sequential decision process: by iterating through vessel slots with a pre-defined sequencer, the task at each step is to select the most suitable container $c\in C$ to place. This requires finding an optimal ordered sequence $\mathbf{P} = \langle p_1, \dots, p_m \rangle$ of $m$
 distinct containers. This problem is subject to numerous constraints, including slot and container availability, bay adjacency rules for 20-foot and 40-foot containers, hatch status, loading sequences for vessel stability, and weight distribution limits.

 Our primary optimization objectives are minimizing shifters and total operation duration. Furthermore, we extend the problem to include crane scheduling, where for $k$ cranes in set $CR$, the goal is to find an optimal sequence of pairs $\langle (p_1, o_1), \dots, (p_m, o_m) \rangle$, where $p_i$ is the container and $o_i\in CR$is the assigned crane.

\section{Related Work}

The NP-hard CSPP is often addressed by decomposing it into Master Bay Planning (MBPP) and Slot Planning (SPP) \cite{wilson2000container}. Early research focused on exact models like 0-1 Linear Programming, which struggled with large-scale instances \cite{AMBROSINO200481}. Subsequent works have refined these approaches using integer programming with progressive constraints \cite{zhu2020integer} and developing strong lower/upper bounds to define feasible solution spaces \cite{cruz2015lower}. Furthermore, some studies have integrated stowage planning with crane scheduling, employing methods such as heuristics and genetic algorithms to solve the combined problem \cite{zheng2010effective,azevedo2018solving}.

The CSPP can be framed as a sequential decision process, making it well-suited for reinforcement learning (RL) solutions, and recent studies have increasingly explored this area. Early work applied Deep Q-Learning, designing reward functions based on plan feasibility and reshuffling frequency \cite{shen2017deep}, with subsequent improvements incorporating Prioritized Experience Replay to handle sparse rewards \cite{xia2020loading}. Other approaches have utilized Monte Carlo Tree Search for planning \cite{zhao2018container}. More recently, Proximal Policy Optimization (PPO) has been adopted to solve subproblems like Master Bay Planning, with reward functions designed to optimize vessel utilization and stability by considering factors such as center of gravity deviation and hatch overstowage \cite{van2023towards,cho2024developing}.

Simulator-based training is essential for RL due to the high cost of real-world data, but discrepancies between simulated and real environments can hinder policy transfer \cite{zhao2020sim}. Research confirms that environment characteristics, such as dynamics and non-stationarity, significantly impact algorithm performance and sample efficiency \cite{henderson2018deep,padakandla2021survey}.
Specific environmental components, including state representation, action space, and reward functions, critically shape agent behavior and learning outcomes \cite{Reda_2020}. Furthermore, evaluating policies across a diverse range of environment variations, rather than on single instances, is crucial for assessing true generalization \cite{jayawardana2022impact}.

\section{Reinforcement Learning Preliminaries}

Reinforcement learning is a machine learning paradigm where an agent learns optimal behavior through trial-and-error interaction with an environment \cite{10.5555/3312046}. This process is typically modeled as a Markov Decision Process (MDP), a tuple $(\mathcal{S}, \mathcal{A}, \mathcal{P}, \mathcal{R})$ representing the state space, action space, state transition dynamics, and reward function, respectively. At each timestep $t$, an agent in state $S_{t}$ executes an action $A_t$, receives a reward $R_t$, and transitions to a new state $S_{t+1}$.

The agent's behavior is defined by its policy $\pi$, which maps states to actions. The goal of RL is to find a policy that maximizes the cumulative discounted reward, not just the immediate reward. To this end, value functions are used to estimate the long-term "goodness" of states or state-action pairs. The state-value function $v_{\pi}(s)$ represents the expected cumulative reward from state $s$, which can be calculated through equation \ref{value_func}, while the action-value function $q_{\pi}(s, a)$ as shown in equation \ref{q-func} represents the expected return after taking action $a$ in state $s$. The critical insight for computing these values is the Bellman equation, which expresses a recursive relationship between a state's value and the value of its potential successor states \cite{bellman1966dynamic}, as shown in equation \ref{bellman}. 
\begin{equation}
v_{\pi}(s) \doteq \mathbb{E}_{\pi}[G_t \mid S_t = s] 
= \mathbb{E}_{\pi} \left[ \sum_{k=0}^{\infty} \gamma^k R_{t+k+1} \,\middle|\, S_t = s \right], \quad \text{for all } s \in \mathcal{S},\label{value_func}
\end{equation}

\begin{equation}
q_{\pi}(s, a) \doteq \mathbb{E}_{\pi}[G_t \mid S_t = s, A_t = a] 
= \mathbb{E}_{\pi} \left[ \sum_{k=0}^{\infty} \gamma^k R_{t+k+1} \,\middle|\, S_t = s, A_t = a \right]\label{q-func}
\end{equation}

\begin{equation}
v_{\pi}(s) = \mathbb{E}_{\pi}[R_{t+1} + \gamma v_{\pi}(S_{t+1}) \mid S_t = s]\label{bellman}
\end{equation}
RL methods can be broadly categorized as value-based or policy-based. Value-based methods, like Q-learning, learn an action-value function 
$Q(s,a)$ and derive a policy by selecting actions that maximize these Q-values. Deep Q-Networks (DQN) extend this to large, continuous state spaces by using neural networks to approximate $Q(s,a)$ \cite{mnih2013playing}.

In contrast, policy-based methods directly learn a parameterized policy $\pi_\theta(a|s)$. They adjust the policy parameters $\theta$ by performing gradient ascent on an objective function to maximize the expected return \cite{williams1992simple}.

\section{Methodology}
\subsection{Environment}
\begin{figure}[htbp]
	\centering
	\subfloat[SPGE]{\includegraphics[width=.49\columnwidth]{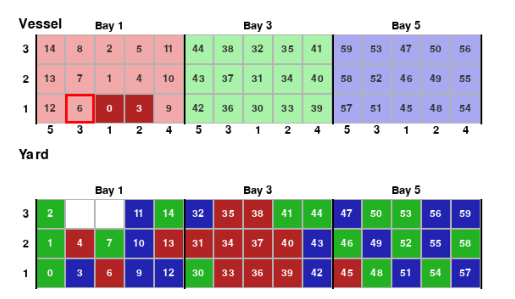}\label{sf:bspge}}
	\subfloat[SPGE-MC]{\includegraphics[width=.49\columnwidth]{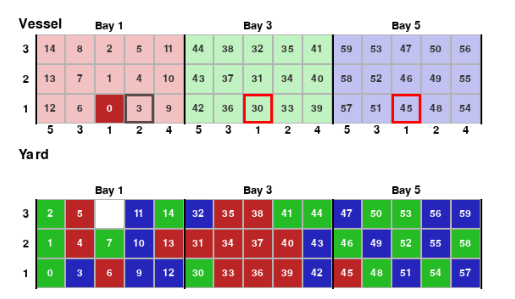}\label{sf:spge-mc}}\\
	\caption{\textit{Visualization of SPGE and SPGE-MC.} Cross-section views of vessel and yard bays. The upper and lower parts of the figure represent the current states of the vessel and yard slots, respectively. Each square represents a container slot, while different colors indicate different groups. Light-color squares in vessel requires to be filled with containers with corresponding group, while white squares in yard are empty slots. Numbers on squares are unique slot IDs, while numbers below and left indicate rows and tiers. In (a)(b), outlined squares are target vessel slots to be filled. The difference is that in (b), multiple sequencers correspond to the number of cranes, and a time mechanism is introduced, which brings crane availability into consideration. A red highlight indicates that the crane associated with the sequencer is idle and the vessel slot can be filled at this step, while gray highlight indicates the slot cannot be filled due to unavailability of the crane.}
        \label{fig:spge}
\end{figure}

We developed the Stowage Planning Gym Environment (SPGE) for our research, an OpenAI Gym-compatible \cite{1606.01540} high-level abstraction of the stowage process. SPGE addresses data scarcity and over-complexity of stowage problem by representing the vessel and yard as configurable cubes of operational slots. Each slot is defined by a state vector containing key attributes: Coordinates (bay, row, tier), Occupancy (1 for present, 0 for empty), and Group (container type). These attributes form the basis for both simplified constraint enforcement and the agent's observation space, which is shown in Table \ref{tab:spge_obs}. With these attributes, SPGE is able to represent complete loading scenarios, such as the one illustrated in Figure \ref{fig:spge}.

In the Basic SPGE environment, the agent's task is sequential container selection. A sequencer dictates the target vessel slot to be filled, and the agent's action is to choose a container from the yard. The observation is a flattened vector of all vessel and yard slot attributes plus the target slot's information. Invalid actions receive a large penalty, while valid actions are rewarded based on the negative number of shifters required. To incorporate crane scheduling, we developed two distinct environment formulations:
\begin{table}[htbp]
\centering
\caption{\textit{Partial observation values of vessel and yard corresponding to Figure \ref{sf:bspge}.}}
\begin{threeparttable}
\begin{tabular}{cccccc}
\multicolumn{6}{c}{\textbf{(a) Vessel State}} \\
\hline
Slot Number & Bay & Row & Tier & Occupancy & Group \\
\hline
0 & 1 & 1 & 1 & 1 & 0 \\
3 & 1 & 2 & 1 & 1 & 0 \\
6 & 1 & 3 & 1 & 0 & 0 \\
30 & 3 & 1 & 1 & 0 & 1 \\
... & ... & ... & ... & ... & ... \\
\hline
\end{tabular}
\vspace{0.5em} 
\begin{tabular}{cccccc}
\multicolumn{6}{c}{\textbf{(b) Yard State}} \\
\hline
Slot Number & Bay & Row & Tier & Occupancy & Group \\
\hline
0 & 1 & 1 & 1 & 1 & 1 \\
1 & 1 & 1 & 2 & 1 & 1 \\
4 & 1 & 2 & 2 & 1 & 0 \\
5 & 1 & 2 & 3 & 0 & -1\tnote{1} \\
... & ... & ... & ... & ... & ... \\
\hline
\end{tabular}
\begin{tablenotes}
\footnotesize
\item[1] Since this yard slot has no container, the container attribute $group$ does not apply.
\end{tablenotes}
\end{threeparttable}
\label{tab:spge_obs}
\end{table}

\begin{table}[htbp]
\centering
\caption{\textit{Crane status related observation values corresponding to Figure \ref{sf:spge-mc}.} The three highlighted squares from left to right correspond to Crane 1, 2, and 3. The size of the state is determined by $n_\text{CR}$.}
\begin{threeparttable}
\begin{tabular}{ccc}
\multicolumn{3}{c}{\textbf{(a) Crane Availability State}} \\
\hline
Crane 1&Crane 2&Crane 3 \\
\hline
100\tnote{1} & 0 & 0 \\
\hline
\end{tabular}
\vspace{0.5em} 
\begin{tabular}{ccc}
\multicolumn{3}{c}{\textbf{(b) Crane Operating State}} \\
\hline
Crane 1&Crane 2&Crane 3 \\
\hline
8 & -1\tnote{2} & -1 \\
\hline
\end{tabular}
\vspace{0.5em} 
\begin{tabular}{ccc}
\multicolumn{3}{c}{\textbf{(c) Crane Sequencer State}} \\
\hline
Crane 1&Crane 2&Crane 3 \\
\hline
3 & 30 & 45 \\
\hline
\end{tabular}
\begin{tablenotes}
\footnotesize
\item[1] Indicates that the crane will be available in 100 seconds.
\item[2] Indicates crane not currently loading a container.
\end{tablenotes}
\end{threeparttable}
\label{tab:spgemc_obs}
\end{table}
\subsubsection{Stowage Planning Multiple Cranes (SPGE-MC)} SPGE-MC models the problem with a single, centralized agent. Its action space is a composite of (container, crane) pairs, allowing it to explicitly control which crane handles which container. Crane availability is managed by a global clock $t$ and a vector $\boldsymbol{\tau}$ that tracks when each crane will become free. The observation space is augmented with crane-specific states, including availability and current task, as shown in Table \ref{tab:spgemc_obs}.

\subsubsection{Stowage Planning Agent Environment Cycle (SPAEC)} \label{sec:spge-mc}

SPAEC models the problem as a multi-agent system compatible with PettingZoo \cite{terry2021pettingzoo} interface, where each crane is an agent controlled by a shared policy. While it uses the same rich observation space and time management logic as SPGE-MC, its action space for each agent is simply container selection. Crane scheduling is handled implicitly: when multiple cranes are available, they are activated in a predefined order (e.g., by bay location), unlike the free choice available in SPGE-MC. This imposes a structural constraint on the decision-making process.
\subsection{Algorithms and Implementation}
For this study, we utilize the Stable-Baselines3 (SB3) framework, a reliable PyTorch-based library, to benchmark five RL algorithms on our custom environments \cite{stable-baselines3}. We selected leading algorithms that support discrete action spaces: the value-based DQN \cite{mnih2013playing} and distributional QR-DQN \cite{dabney2018distributional}, alongside the actor-critic methods A2C \cite{mnih2016asynchronous}, TRPO \cite{schulman2015trust}, and PPO \cite{schulman2017proximal}.

To enhance training efficiency in our complex, constrained environments, we integrated action masking into all algorithms. This technique filters out invalid actions at the policy or Q-network level, preventing the agent from exploring futile actions. As demonstrated by \cite{huang2020closer}, action masking dramatically accelerates convergence and improves scalability compared to naïve penalty-based approaches.

\subsection{Experimental Setup}
We designed eight scenarios with progressively increasing complexity across our SPGE, SPGE-MC, and SPAEC environments, as detailed in Table \ref{tab:scenarios}. These scenarios vary in vessel/yard size, container count (45 to 200), type diversity (3 to 8), and the number of cranes (1, 3, or 5). Scenarios 1-5, featuring a single crane, were used to benchmark algorithms in the Basic SPGE environment. Scenarios 6-8, which involve multiple cranes, were used for comparative analysis between the single-agent (SPGE-MC) and multi-agent (SPAEC) formulations.
\begin{table}[htbp]
\centering
\caption{\textit{Details of the Experimental Scenarios. }Sizes are Bays $\times$ Rows $\times$ Tiers.}
\label{tab:scenarios}
\begin{tabular}{l c c c c c}
\hline
Scen. & \begin{tabular}{@{}c@{}}Vessel Sizes (Slots)\end{tabular} & \begin{tabular}{@{}c@{}}Yard Sizes (Slots)\end{tabular} & \begin{tabular}{@{}c@{}}Cont. to Stow\end{tabular} & \begin{tabular}{@{}c@{}}Cont. Types\end{tabular} & Cranes \\
\hline
1 & 3$\times$5$\times$3 (45) & 3$\times$5$\times$3 (45) & 45 & 3 & 1 \\ 
2 & 3$\times$5$\times$3 (45) & 3$\times$5$\times$3 (45) & 45 & 8 & 1 \\ 
3 & 3$\times$5$\times$3 (45) & \begin{tabular}{@{}c@{}}8$\times$5$\times$5  (200)\end{tabular} & 45 & 8 & 1 \\ 
4 & \begin{tabular}{@{}c@{}}8$\times$5$\times$5 (200)\end{tabular} & 3$\times$5$\times$3 (45) & 45 & 8 & 1 \\ 
5 & \begin{tabular}{@{}c@{}}8$\times$5$\times$5 (200)\end{tabular} & \begin{tabular}{@{}c@{}}8$\times$5$\times$5 (200)\end{tabular} & 200 & 8 & 1 \\
6 & 3$\times$5$\times$3 (45) & \begin{tabular}{@{}c@{}}8$\times$5$\times$5  (200)\end{tabular} & 45 & 8 & 3 \\ 
7 & \begin{tabular}{@{}c@{}}8$\times$5$\times$5 (200)\end{tabular} & \begin{tabular}{@{}c@{}}8$\times$5$\times$5 (200)\end{tabular} & 200 & 8 & 3 \\
8 & \begin{tabular}{@{}c@{}}8$\times$5$\times$5 (200)\end{tabular} & \begin{tabular}{@{}c@{}}8$\times$5$\times$5 (200)\end{tabular} & 200 & 8 & 5 \\
\hline
\end{tabular}
\end{table}

To ensure statistical robustness, we performed 10 independent training repetitions for the SPGE experiments, and 30 repetitions for the more complex SPGE-MC and SPAEC experiments.

\subsection{Results and Discussion}

This section presents the comparative performance of five RL algorithms across our custom environments.

\begin{figure}[!h]
    \centering


    \subfloat[Scenario 1]{%
        \includegraphics[width=0.33\columnwidth, keepaspectratio]{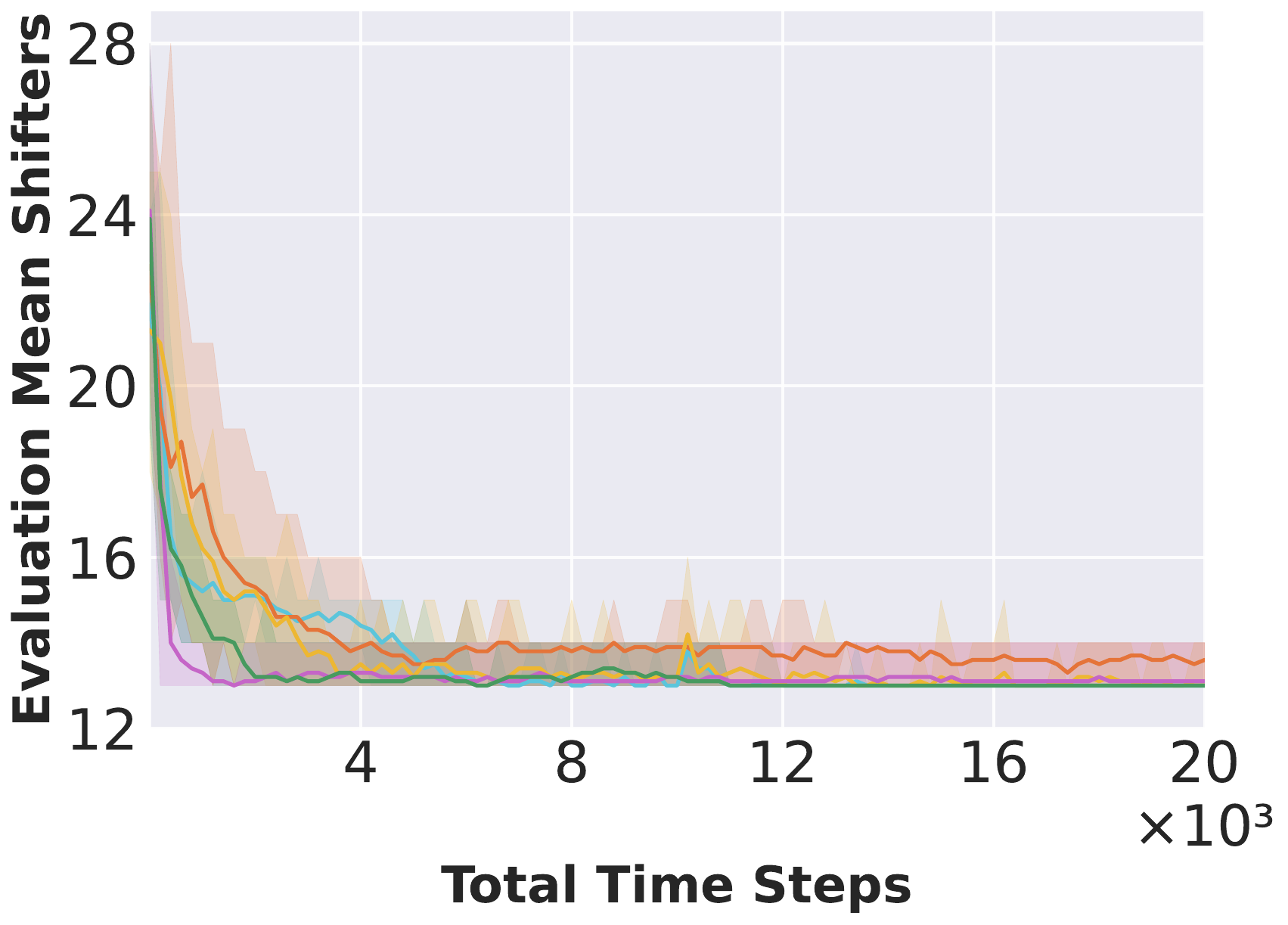}
        \label{sf:s1_eval_shifters_32}
    }
    \hfill
    \subfloat[Scenario 2]{%
        \includegraphics[width=0.32\columnwidth, keepaspectratio]{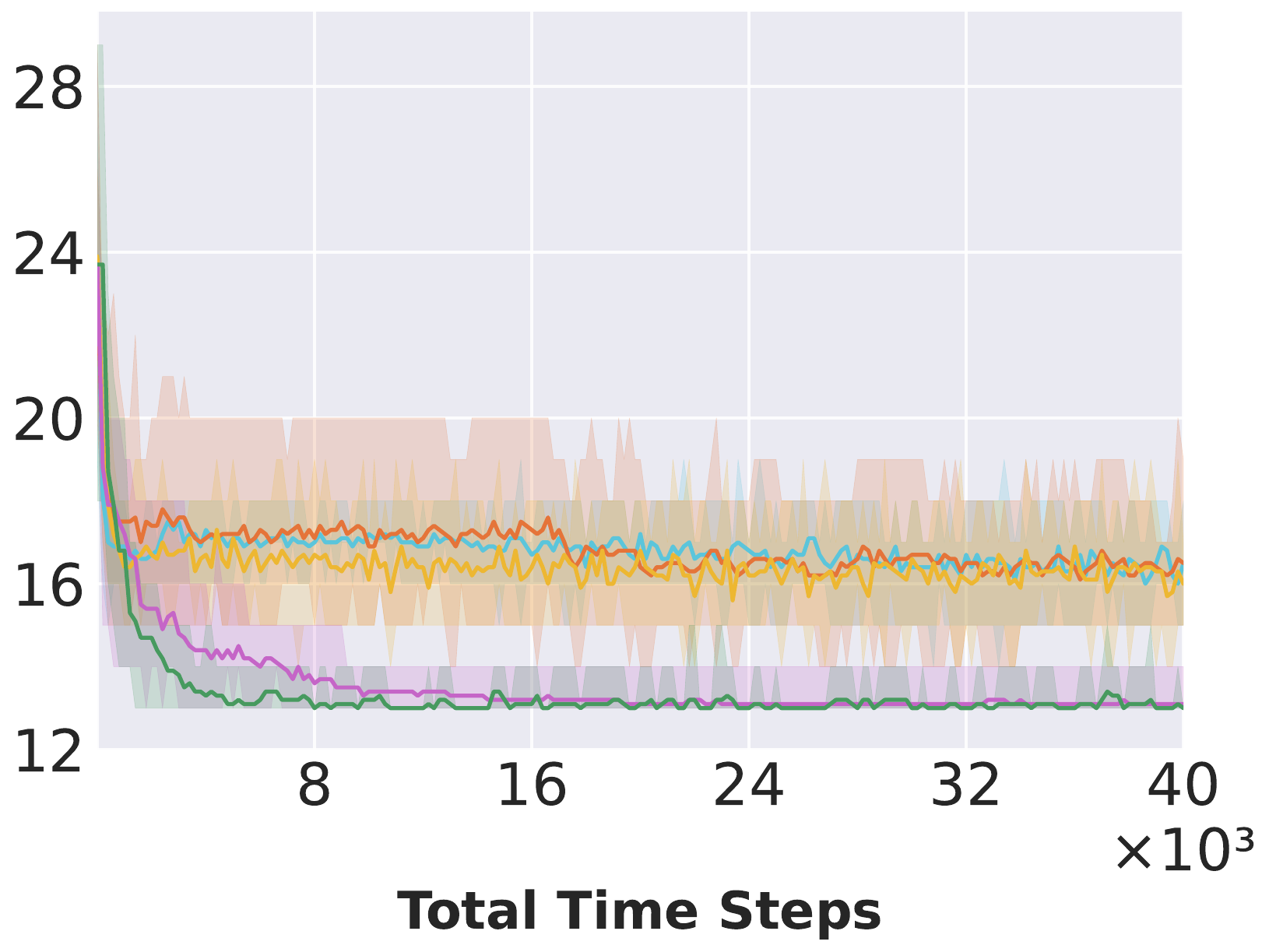}
        \label{sf:s2_eval_shifters_32}
    }
    \hfill
    \subfloat[Scenario 3]{%
        \includegraphics[width=0.32\columnwidth,  keepaspectratio]{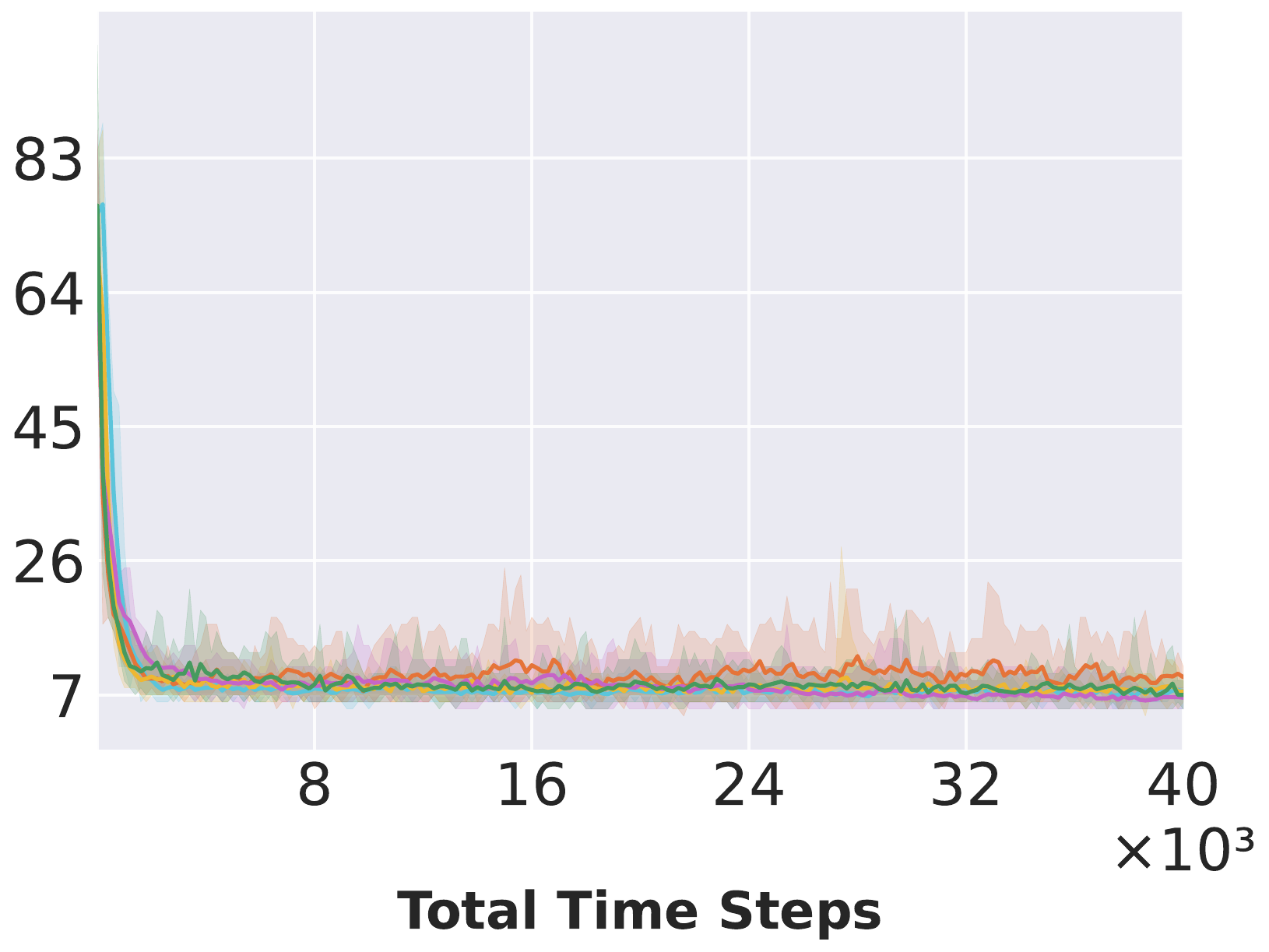}
        \label{sf:s3_eval_shifters_32}
    }
    \\[5pt] 
    \subfloat[Scenario 4]{%
        \includegraphics[width=0.33\columnwidth,  keepaspectratio]{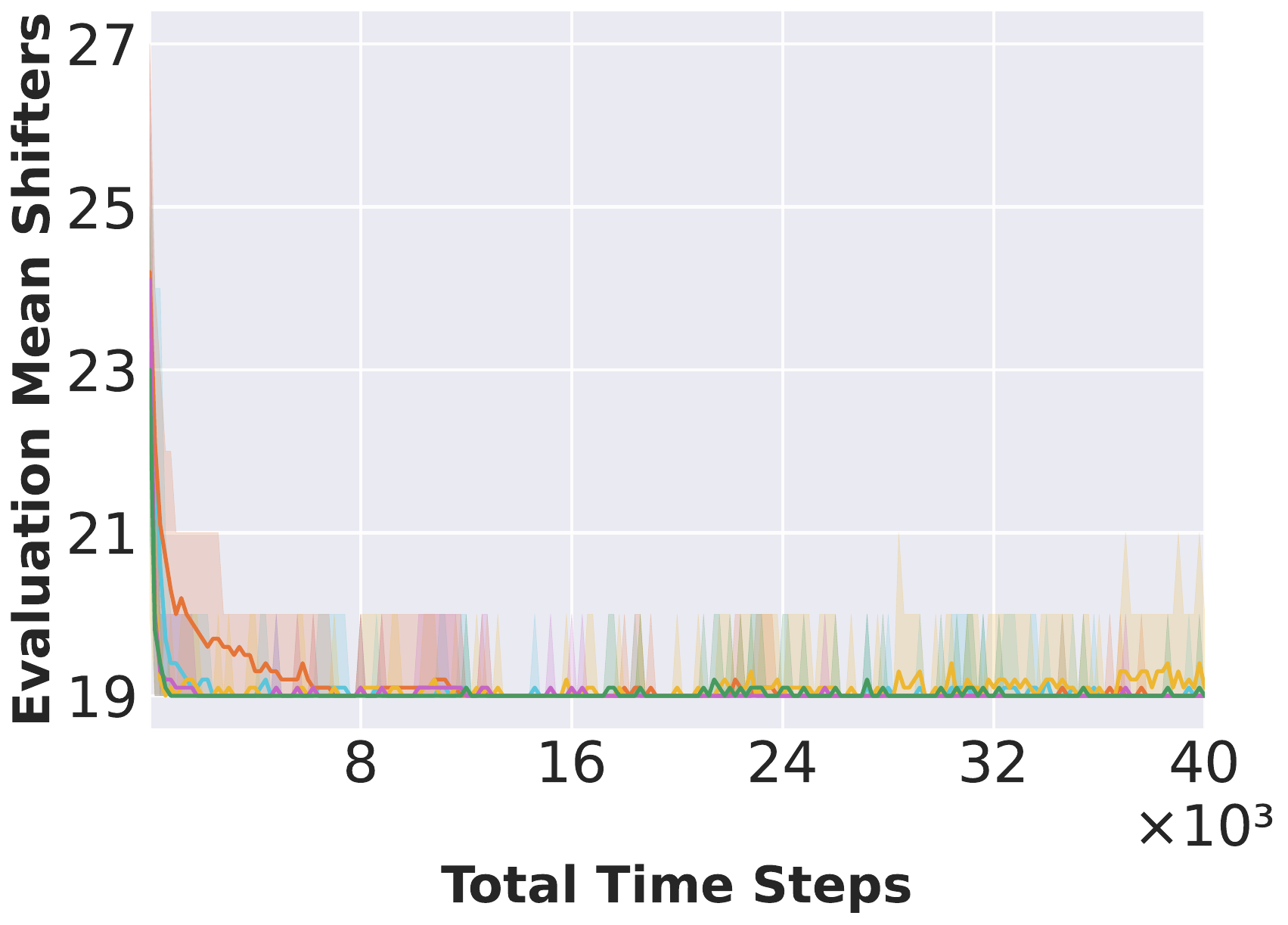}
        \label{sf:s4_eval_shifters_32}
    }
    \subfloat[Scenario 5]{%
        \includegraphics[width=0.32\columnwidth,  keepaspectratio]{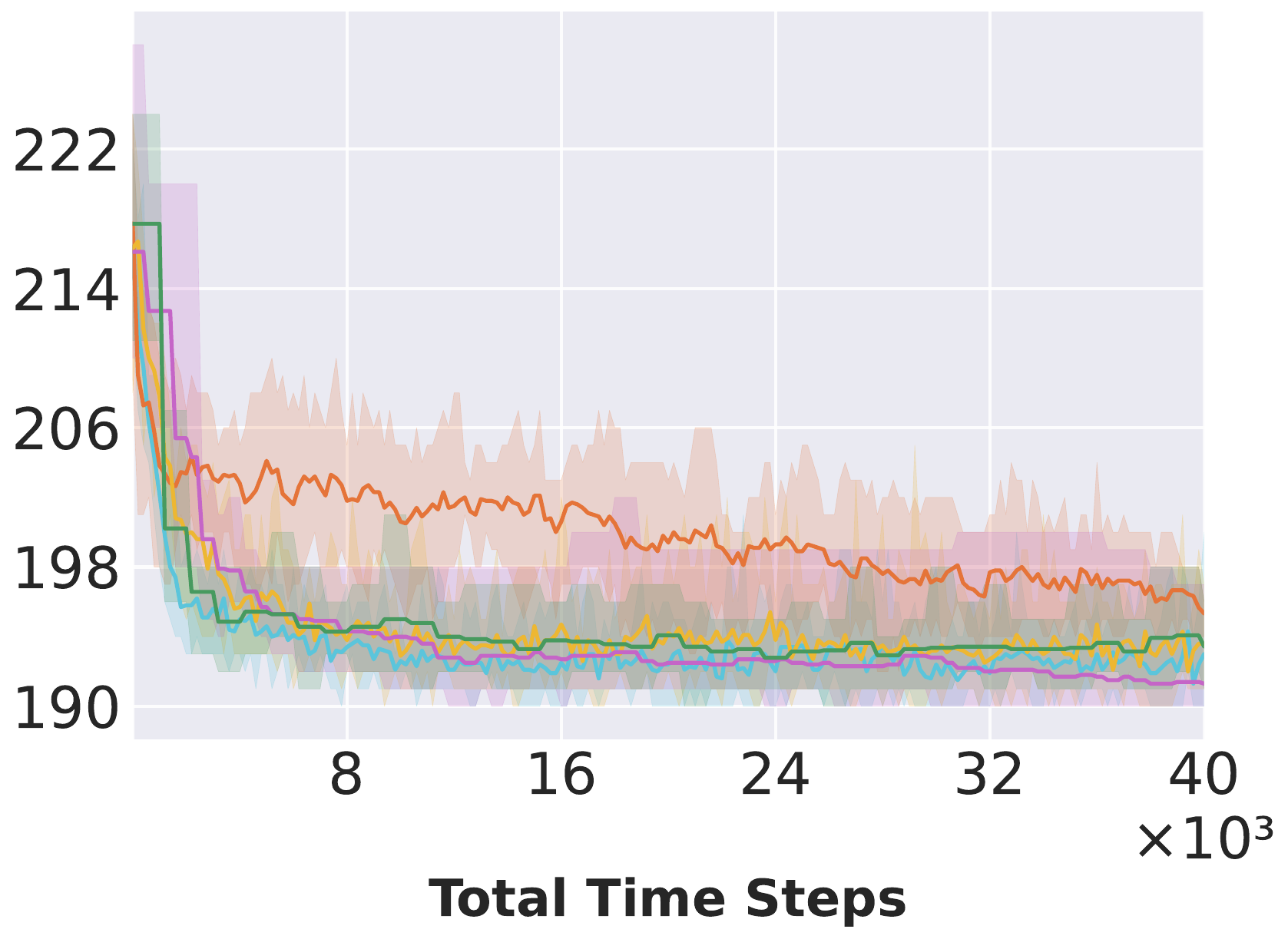}
        \label{sf:s5_eval_shifters_32}
    }
\caption*{%
        \textcolor[RGB]{91, 197, 219}{\rule[0.5ex]{1em}{2pt}} QR-DQN \quad
        \textcolor[RGB]{237, 183, 50}{\rule[0.5ex]{1em}{2pt}} DQN \quad
        \textcolor[RGB]{229, 116, 57}{\rule[0.5ex]{1em}{2pt}} A2C \quad
        \textcolor[RGB]{71, 154, 95}{\rule[0.5ex]{1em}{2pt}} PPO \quad
        \textcolor[RGB]{197, 101, 199}{\rule[0.5ex]{1em}{2pt}} TRPO
    }
    \caption{\textit{Evaluation performance (number of shifters vs. total timesteps) for five algorithms across different scenarios in SPGE.} Subfigures (a)-(e) correspond to Scenarios 1-5. Lower shifter counts indicate better performance. Data based on 10 evaluation trials and 10 training repetitions.}
    \label{fig:evaluation_shifters_all_scenarios_32}

\end{figure}
Figures \ref{fig:evaluation_shifters_all_scenarios_32} show the number of shifters during evaluation for five algorithms under each scenario in SPGE. In simple, single-crane scenarios, all algorithms performed comparably. However, as problem complexity increased (e.g., more container types or larger scale), performance diverged significantly. A2C consistently yielded the poorest results. Value-based methods (DQN, QR-DQN) struggled with increased combinatorial complexity in small-scale problems, converging to suboptimal policies similar to A2C. In contrast, TRPO and PPO demonstrated robust performance, with TRPO achieving the best results in the most complex scenarios, such as maintaining a low shifter count of around 193 in the 200-container case (Scenario 5). 

This may be related to the accuracy of the critic network in estimating state values: in training A2C, we adopted the original paper's choice of 5 for the n-step parameter, but in complex problems, the true value of a single action may not be reflected solely by the immediate shifter, whose long-term impact might only emerge far beyond 5 steps. This may lead to significant bias in the advantage estimation based on 5-step returns. Furthermore, unlike PPO and TRPO, A2C does not use Generalized Advantage Estimation (GAE), which allows for a trade-off between bias and variance in value estimation, and thus cannot benefit from the improved credit assignment that GAE provides.

\begin{figure}[htbp]
    \centering

    \subfloat[SPGE-MC: Scenario 6]{%
        \includegraphics[width=0.33\columnwidth,  keepaspectratio]{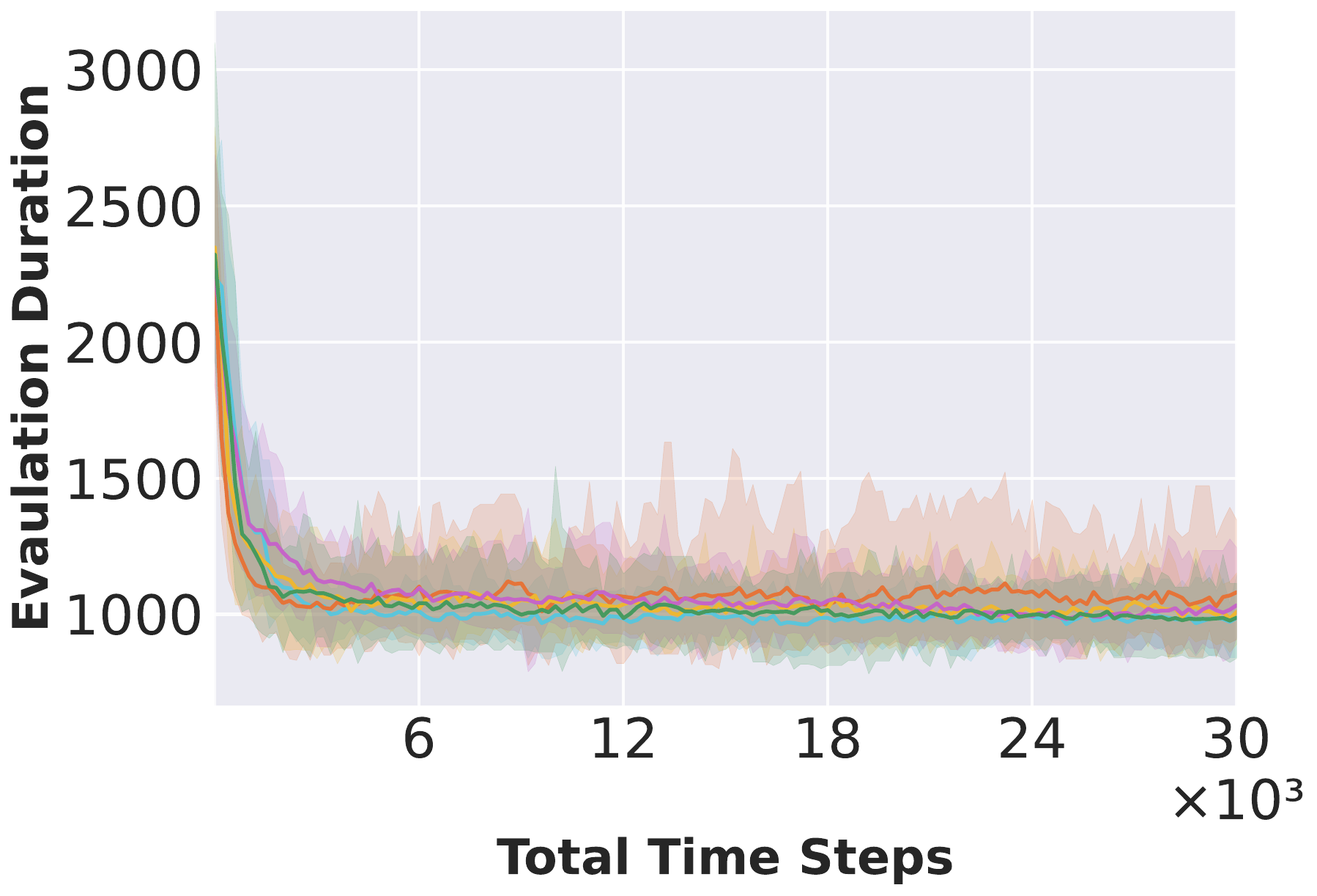}
        \label{sf:spgemc_s1_optime}
    }
    \hfill 
    \subfloat[SPGE-MC: Scenario 7]{%
        \includegraphics[width=0.32\columnwidth,  keepaspectratio]{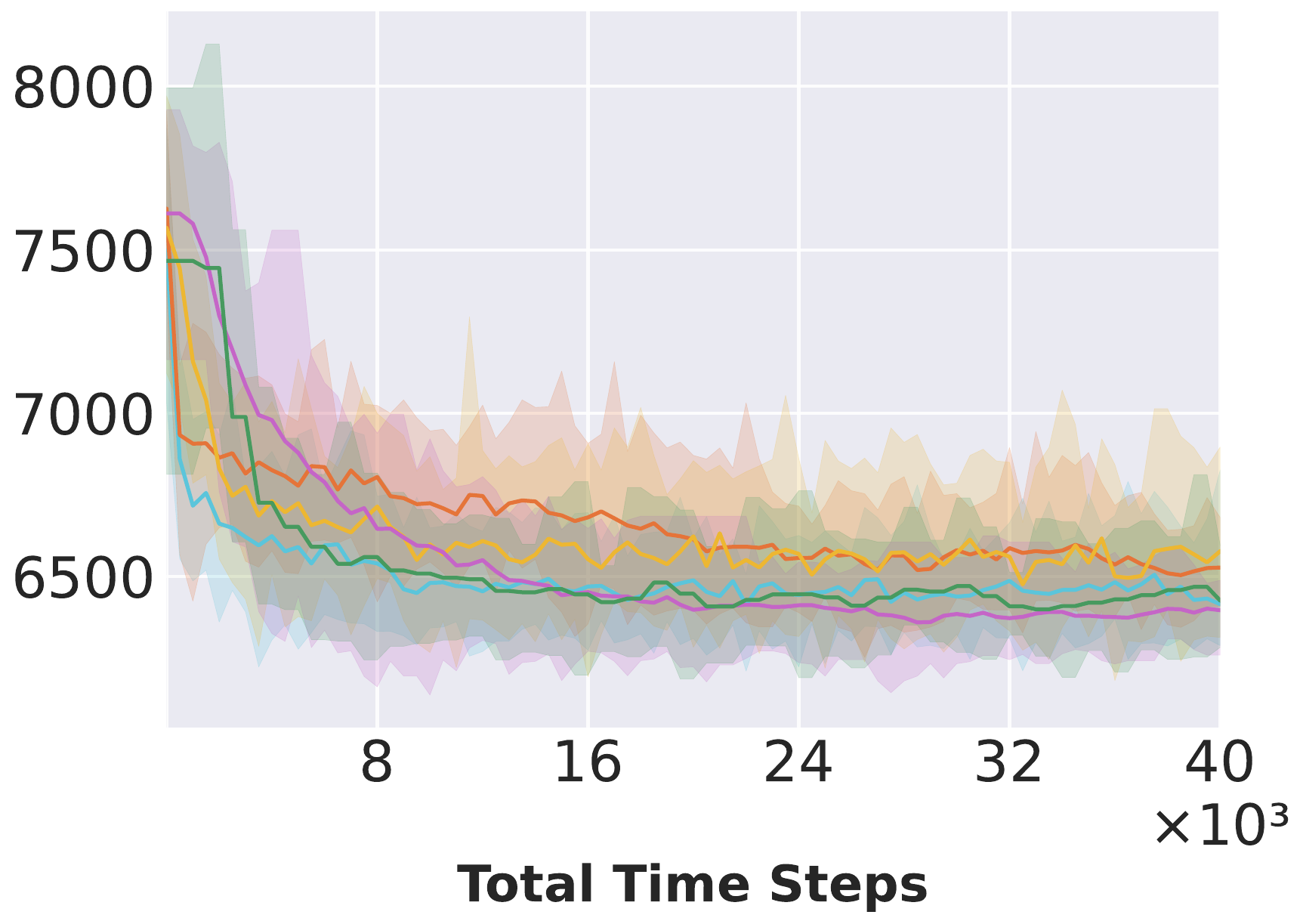}
        \label{sf:spgemc_s2_optime}
    }
    \hfill
    \subfloat[SPGE-MC: Scenario 8]{%
        \includegraphics[width=0.32\columnwidth,  keepaspectratio]{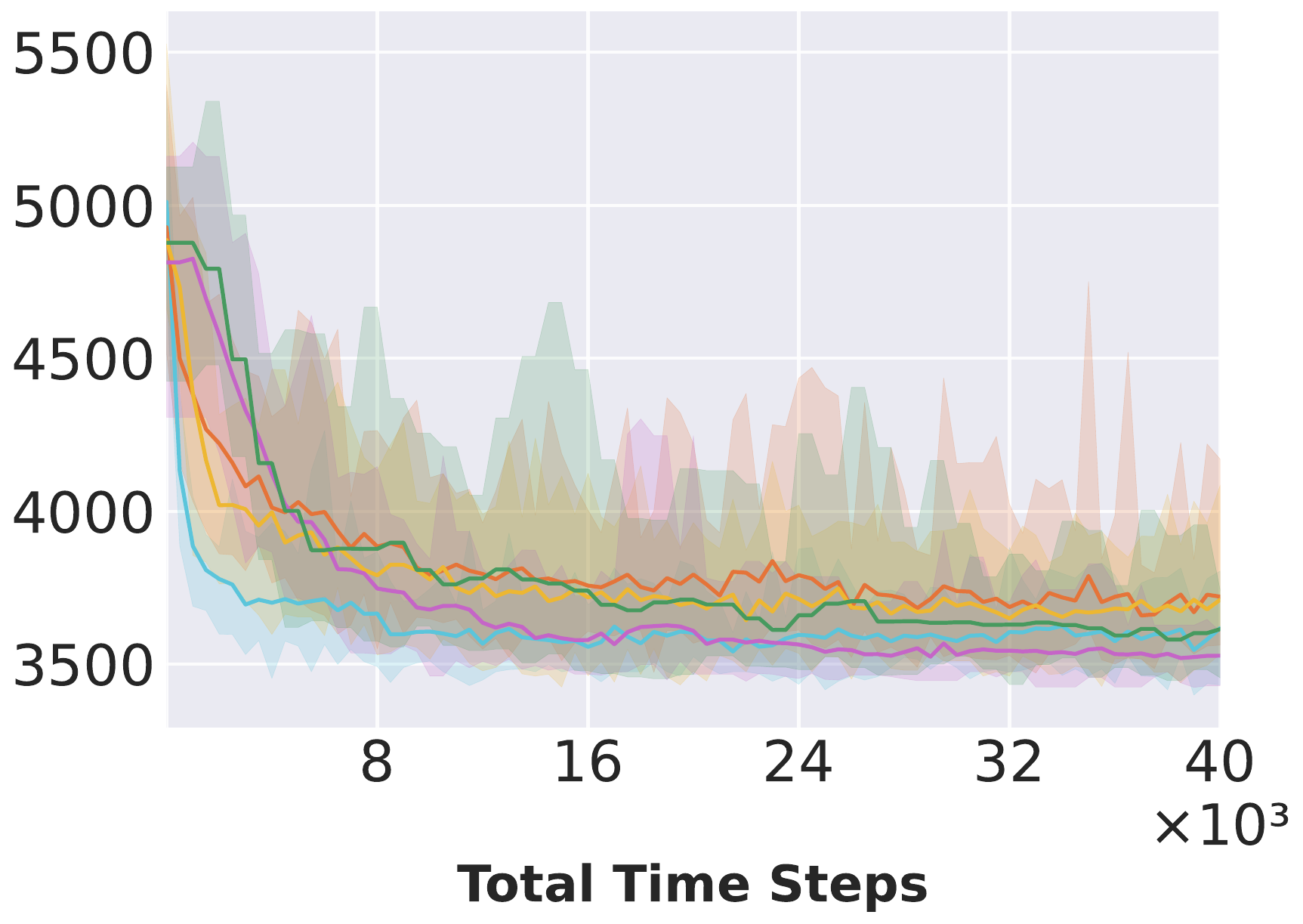}
        \label{sf:spgemc_s3_optime}
    }\\
        \subfloat[SPAEC: Scenario 6 ]{%
        \includegraphics[width=0.32\columnwidth,  keepaspectratio]{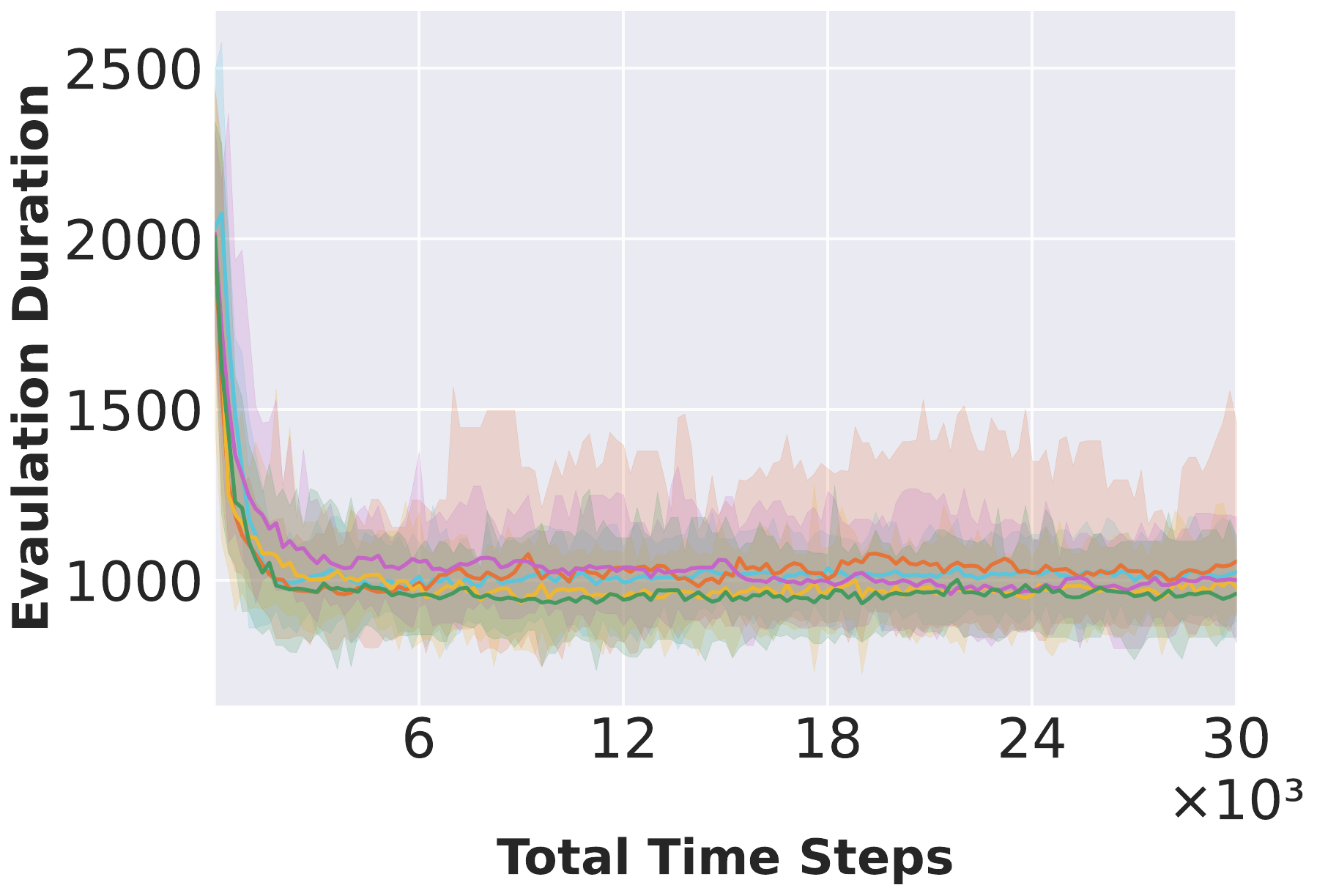}
        \label{sf:spaec_s1_optime}
    }
    \hfill 
    \subfloat[SPAEC: Scenario 7]{%
        \includegraphics[width=0.32\columnwidth,  keepaspectratio]{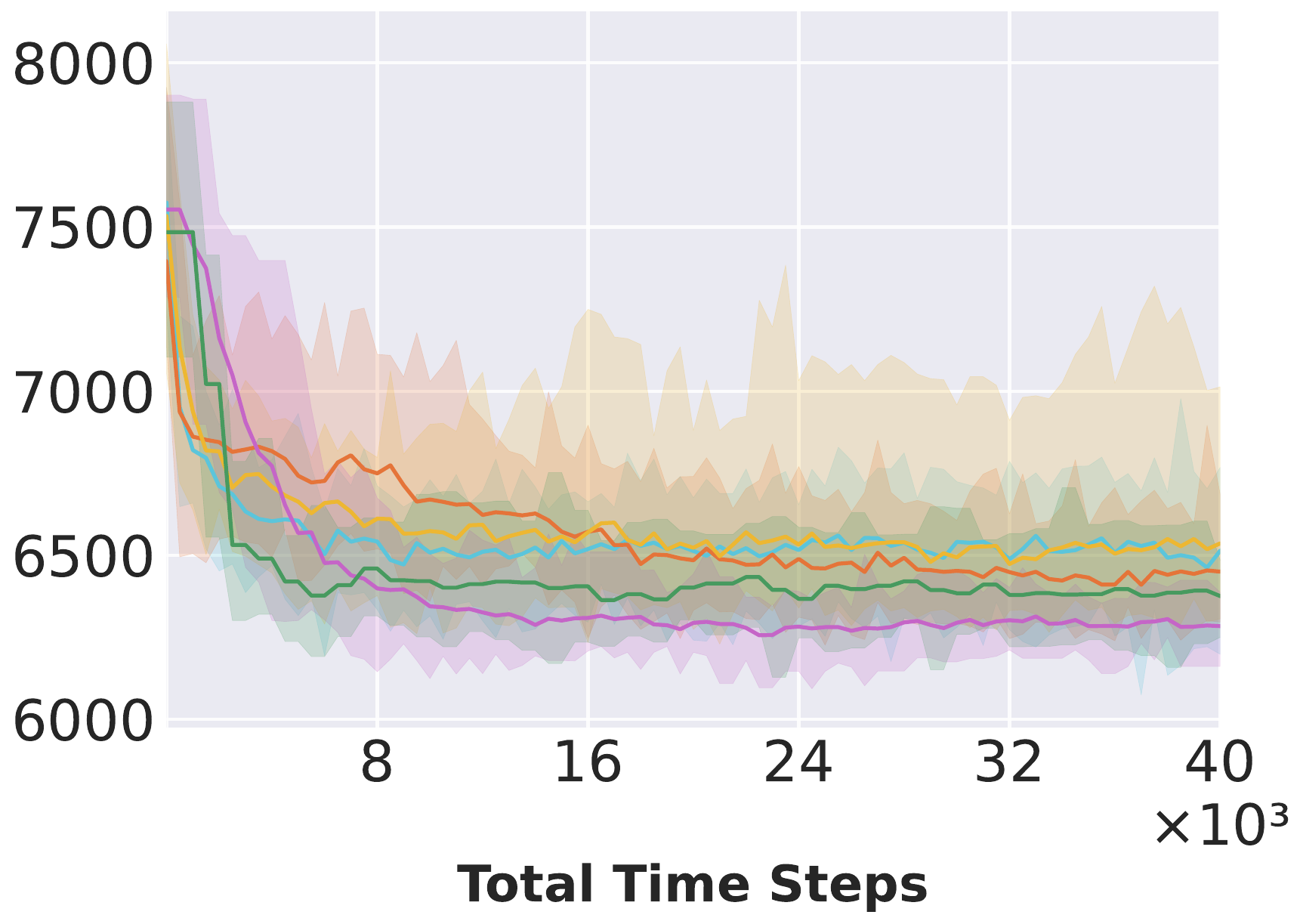}
        \label{sf:spaec_s2_optime}
    }
    \hfill
    \subfloat[SPAEC: Scenario 8]{%
        \includegraphics[width=0.32\columnwidth,  keepaspectratio]{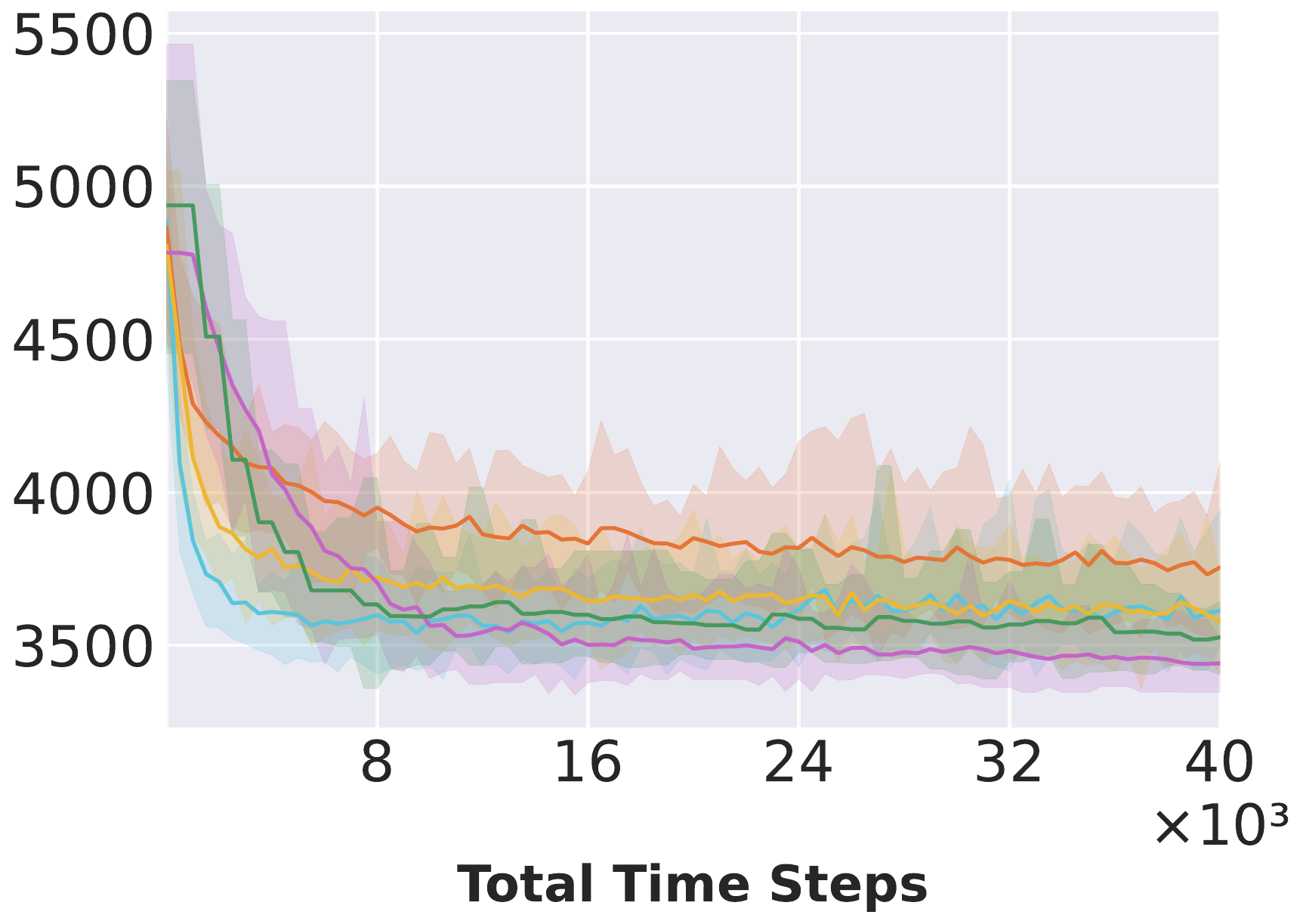}
        \label{sf:spaec_s3_optime}
    }\\
        \subfloat[SPGE-MC: Scenario 6]{%
        \includegraphics[width=0.33\columnwidth,  keepaspectratio]{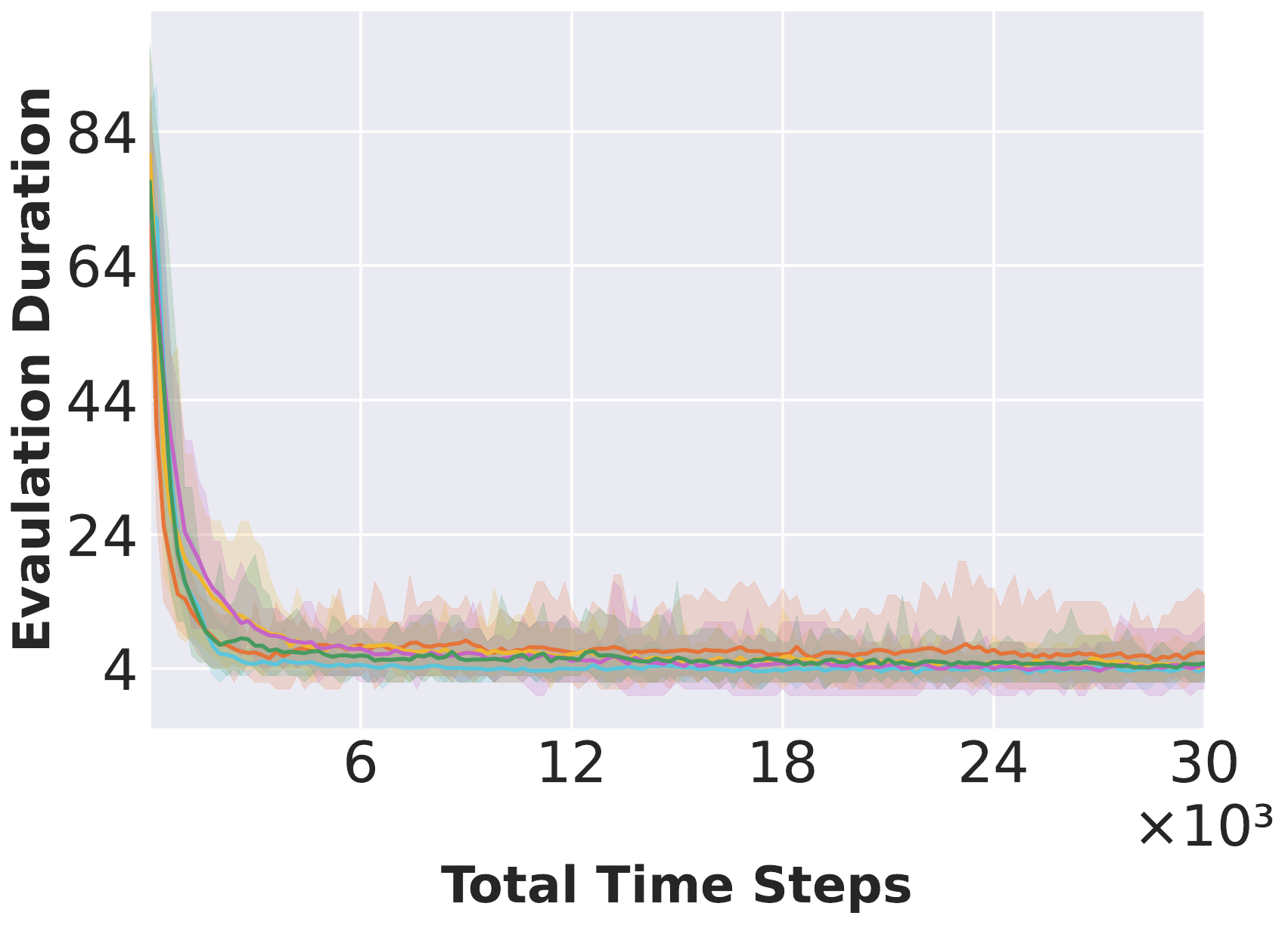}
        \label{sf:spgemc_s1_shifters}
    }
    \hfill
    \subfloat[SPGE-MC: Scenario 7]{%
        \includegraphics[width=0.32\columnwidth,  keepaspectratio]{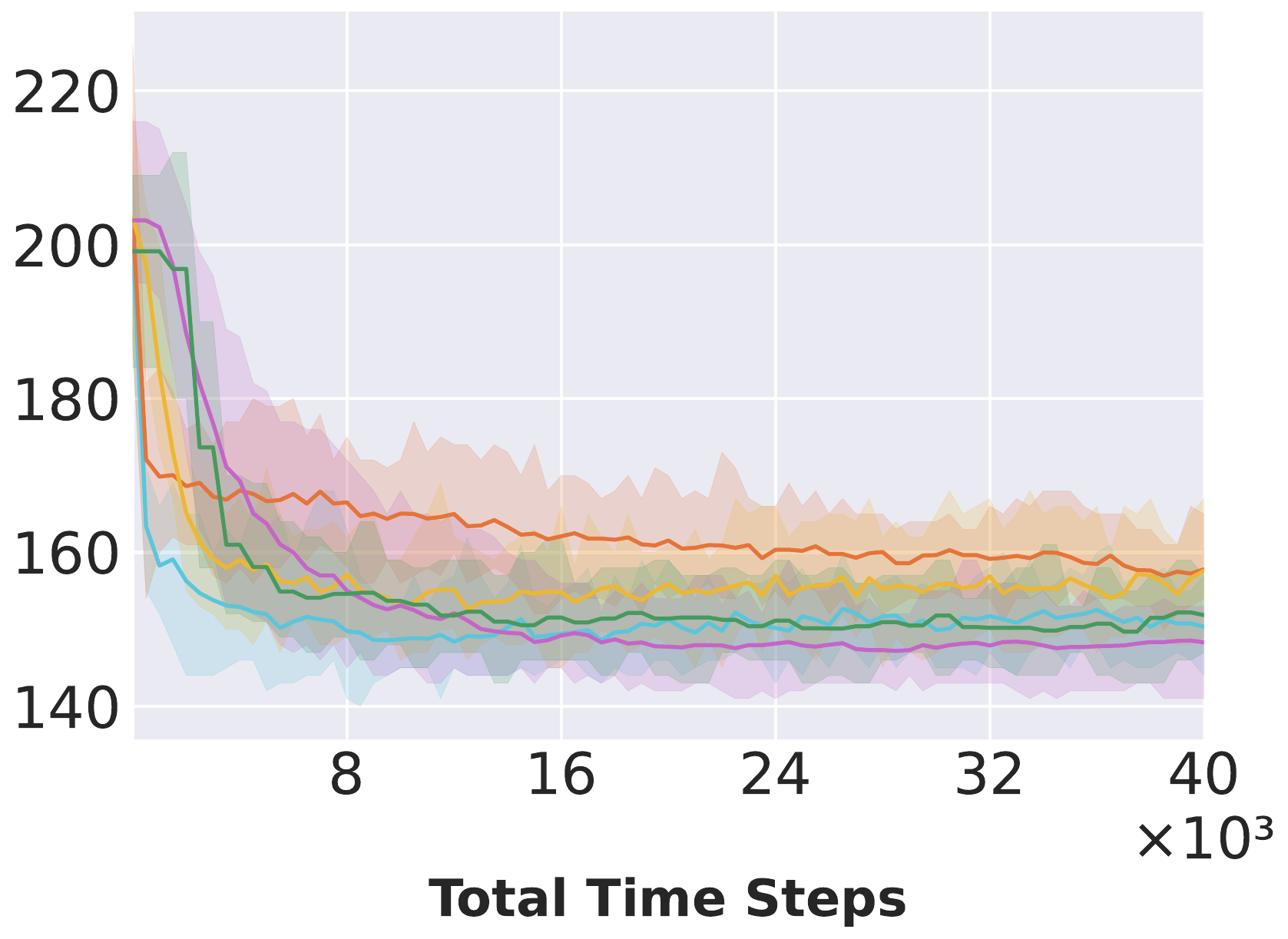}
        \label{sf:spgemc_s2_shifters}
    }
    \hfill
    \subfloat[SPGE-MC: Scenario 8]{%
        \includegraphics[width=0.32\columnwidth,  keepaspectratio]{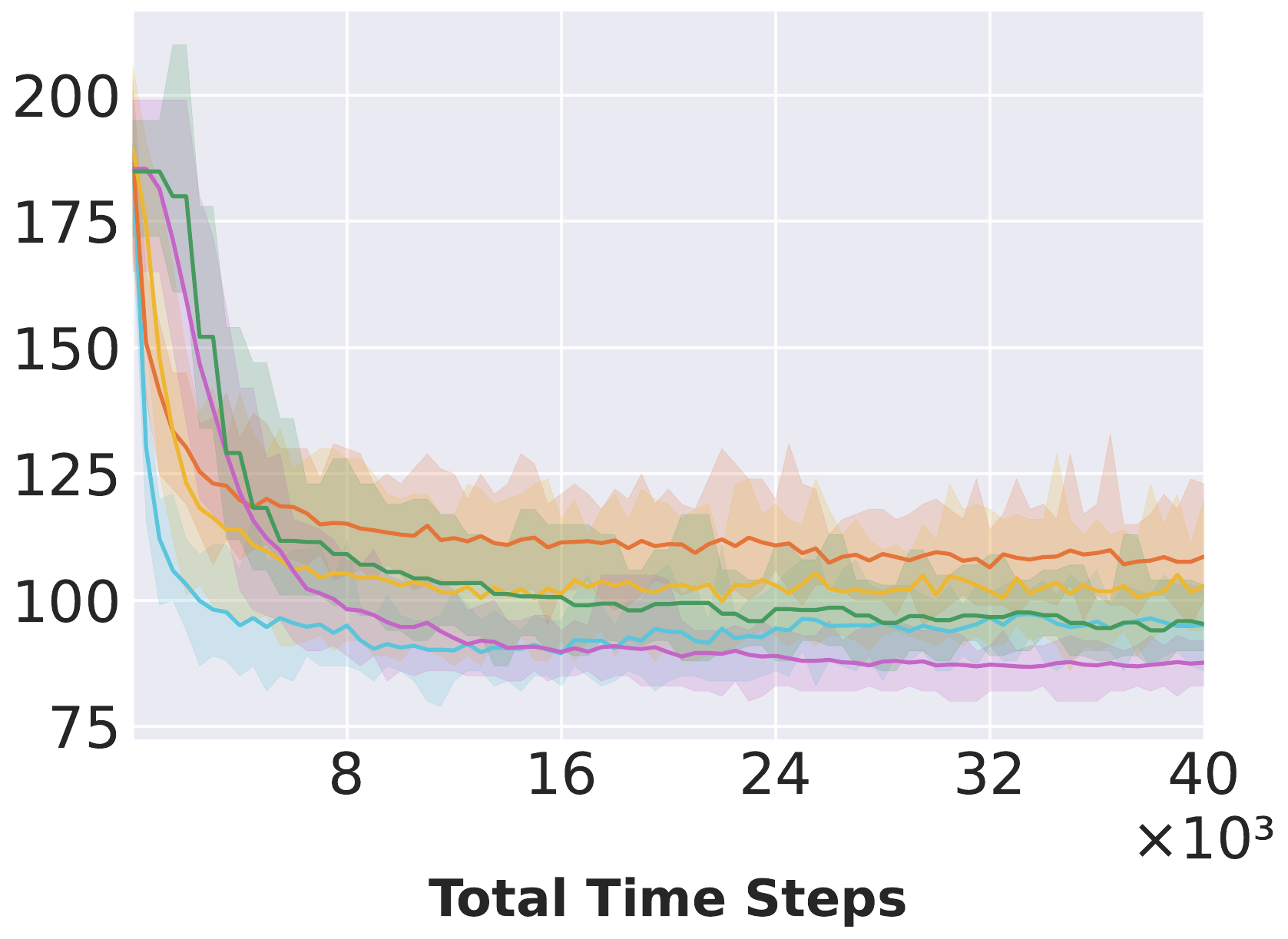}
        \label{sf:spgemc_s3_shifters}
    }\\
        \subfloat[SPAEC: Scenario 6]{%
        \includegraphics[width=0.32\columnwidth,  keepaspectratio]{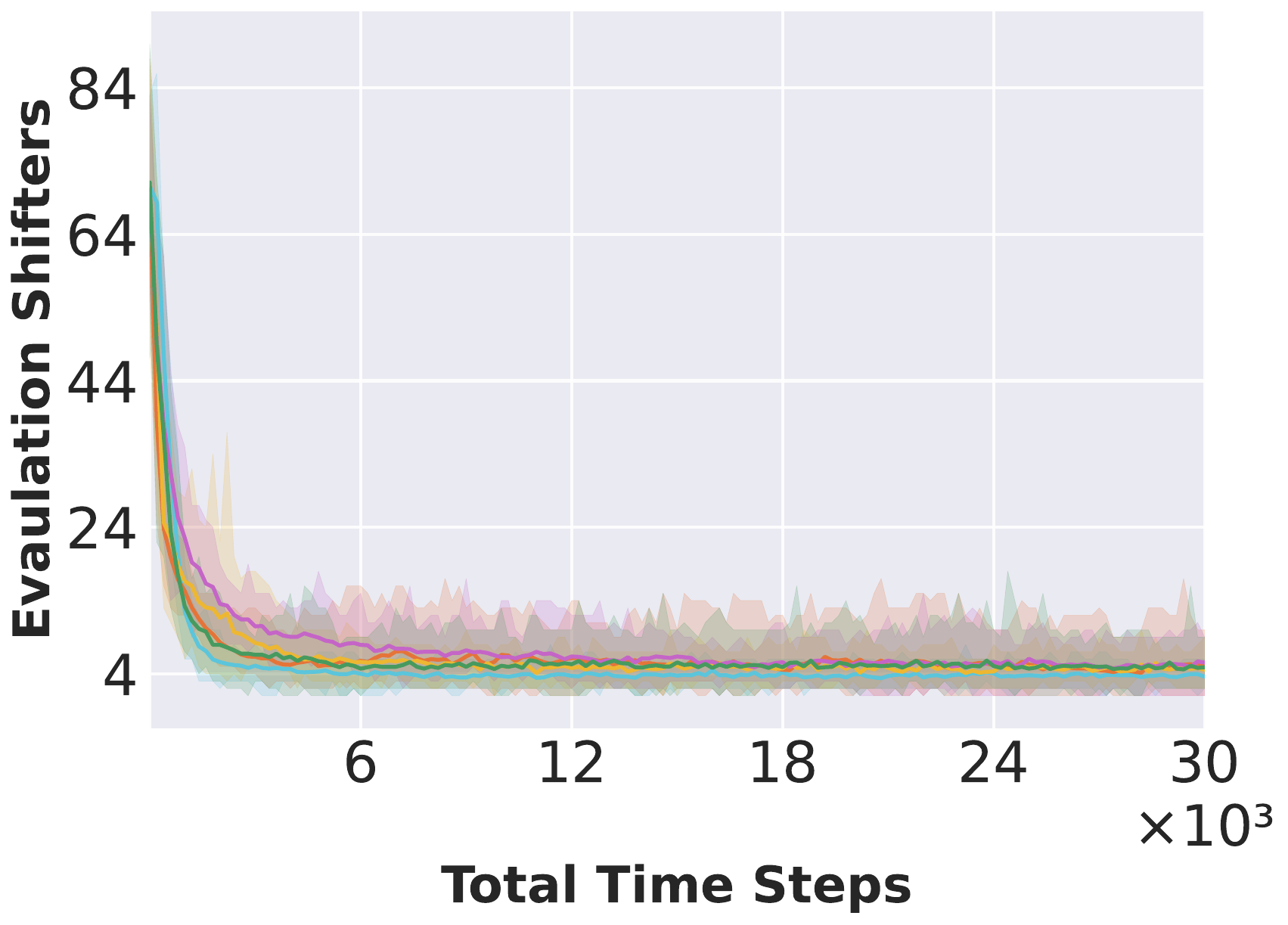}
        \label{sf:spaec_s1_shifters}
    }
    \hfill
    \subfloat[SPAEC: Scenario 7]{%
        \includegraphics[width=0.32\columnwidth,  keepaspectratio]{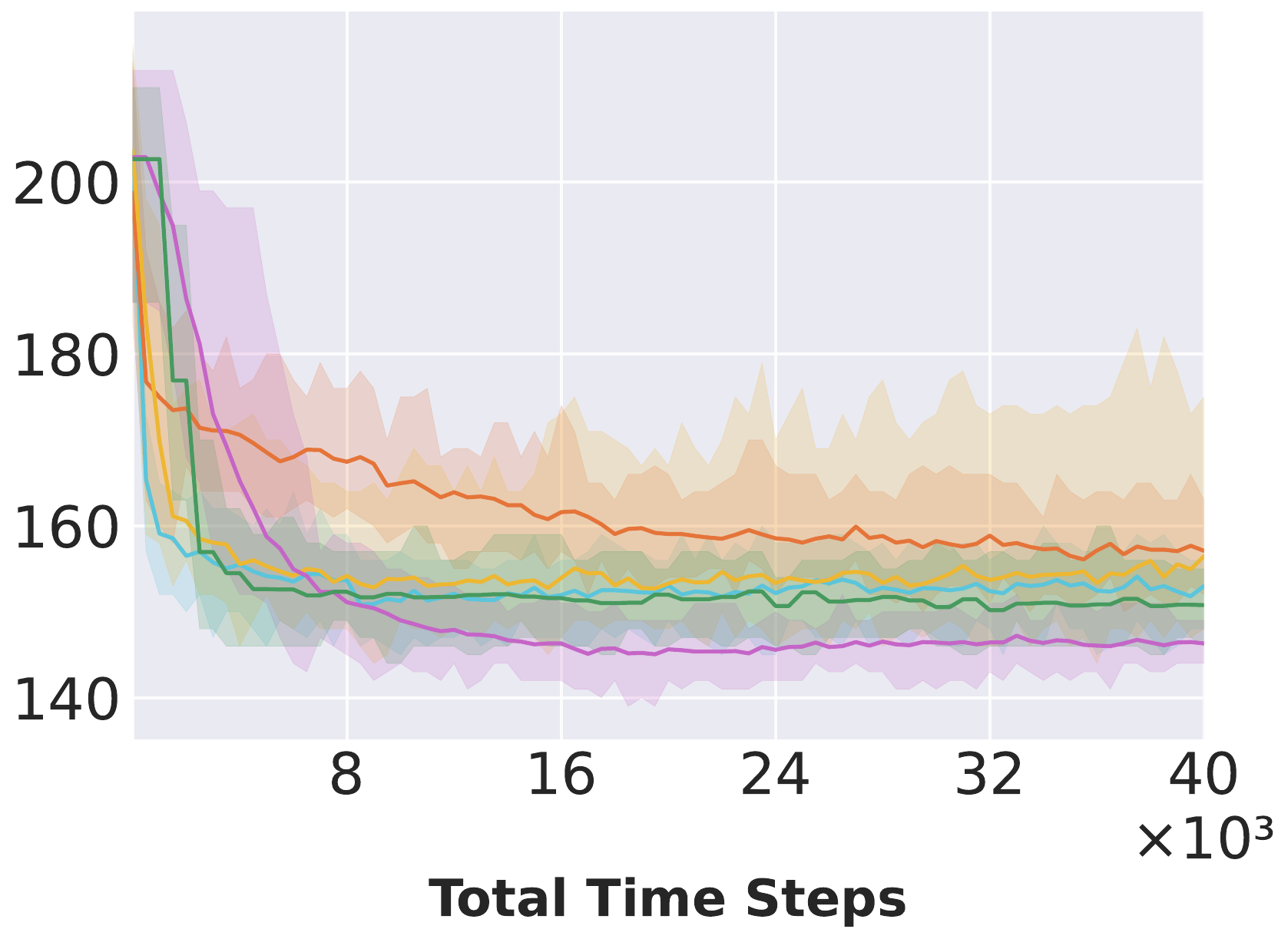}
        \label{sf:spaec_s2_shifters}
    }
    \hfill
    \subfloat[SPAEC: Scenario 8]{%
        \includegraphics[width=0.32\columnwidth,  keepaspectratio]{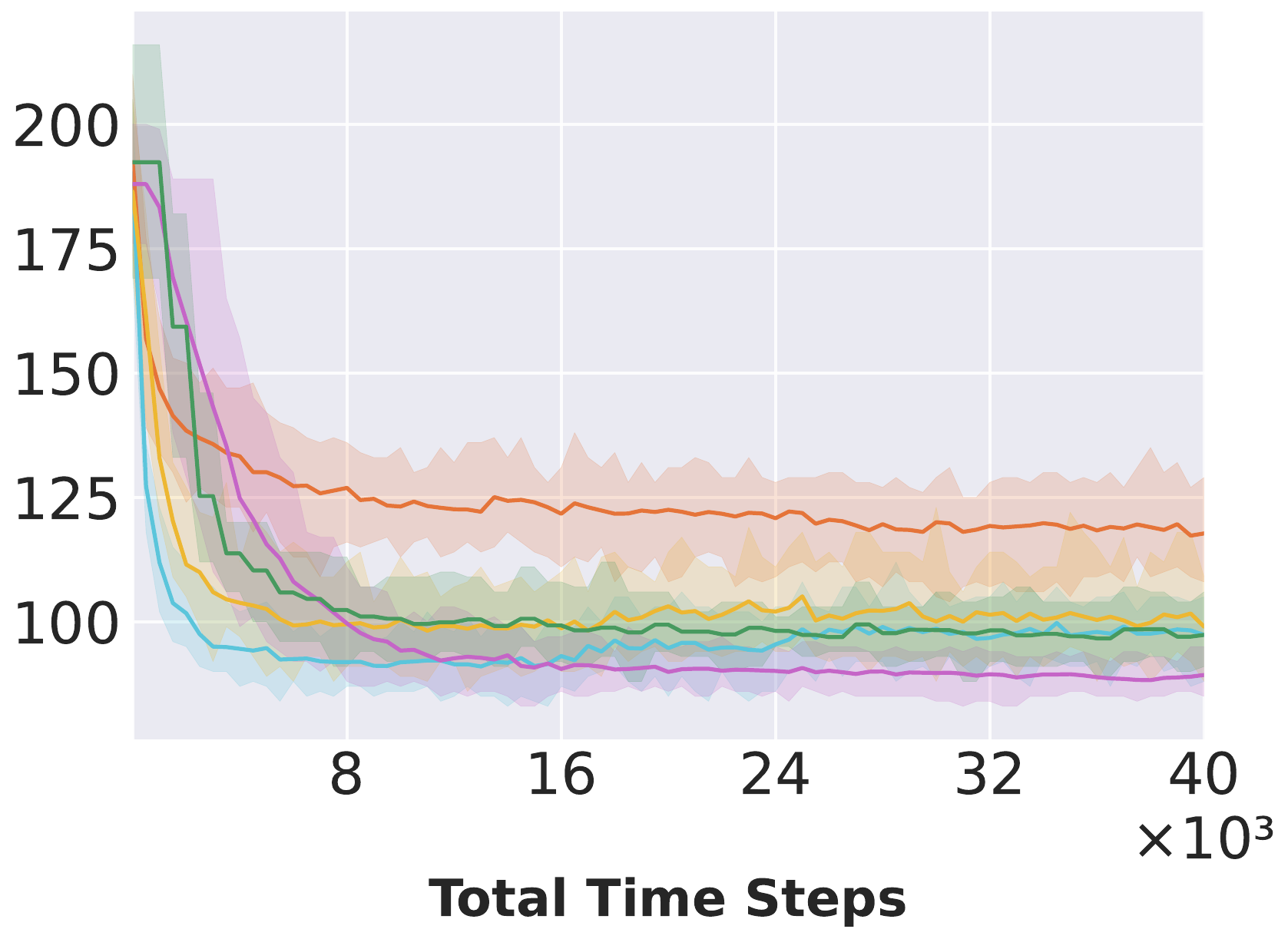}
        \label{sf:spaec_s3_shifters}
    }
        \caption*{%
        \textcolor[RGB]{91, 197, 219}{\rule[0.5ex]{1em}{2pt}} QR-DQN \quad
        \textcolor[RGB]{237, 183, 50}{\rule[0.5ex]{1em}{2pt}} DQN \quad
        \textcolor[RGB]{229, 116, 57}{\rule[0.5ex]{1em}{2pt}} A2C \quad
        \textcolor[RGB]{71, 154, 95}{\rule[0.5ex]{1em}{2pt}} PPO \quad
        \textcolor[RGB]{197, 101, 199}{\rule[0.5ex]{1em}{2pt}} TRPO
    }
    \caption{\textit{Average episode operation time and shifters(vs. total timesteps) for various algorithms in the SPGE-MC environment.}
    (a) Scenario 6: 45 containers, 3 cranes (eval. every 200 timesteps).
    (b) Scenario 7: 200 containers, 3 cranes (eval. every 500 timesteps, 30 training runs).
    (c) Scenario 8: 200 containers, 5 cranes (eval. every 500 timesteps, 30 training runs).
    Lower operation time and shifters indicate better performance.
    }
    \label{fig:spgemc_optime_curves} 
\end{figure}

\begin{table}[htbp]
\centering
\caption{\textit{Comparison of Final Evaluation Metrics (Shifters and Operation Time) across SPGE-MC and SPAEC Environments.} Lower values are better for both KPIs. Values in parentheses indicate standard deviation. Best value per scenario/KPI/environment in \textbf{\textcolor{mygreen}{green}}, worst in \textcolor{myred}{red}. Difference (SMC-SPAEC): \textcolor{myblue}{Blue} if SPGE-MC is significantly better (negative diff), \textcolor{mypurple}{Purple} if SPAEC is significantly better (positive diff). Only differences that are statistically significant based on t-tests are color-coded.}
\label{tab:final_kpi_comparison_colored_full}
\footnotesize
\setlength{\tabcolsep}{3pt}
\begin{tabular}{@{}l l l S[table-format=-3.1,table-text-alignment=center] S[table-format=-4.1,table-text-alignment=center] S[table-format=-3.1,table-text-alignment=center]@{}}
\toprule
\multirow{2}{*}{\textbf{Algorithm}} & \multirow{2}{*}{\textbf{Scenario}} & \multirow{2}{*}{\textbf{KPI}} & {\textbf{SPGE-MC}} & {\textbf{SPAEC}} & {\textbf{Diff.}} \\
 &  &  & {\textbf{Value}} & {\textbf{Value}} & {\textbf{(SMC-SPAEC)}} \\
\midrule
\multirow{6}{*}{TRPO}   & \multirow{2}{*}{Scenario 6} & Shifters & {4.2 (2.25)} & {4.8 (2.82)} & {-0.5} \\ 
                        &                             & Op. Time & {1025.9 (85.95)} & {983.8 (86.10)} & {42.1} \\
\cmidrule(lr){2-6}
                        & \multirow{2}{*}{Scenario 7} & Shifters & {\bestval{148.2 (2.82)}}& {\bestval{146.3 (1.86)}} & \diffcolor{1.9} \\ 
                        &                             & Op. Time & \bestval{6401.2 (52.76)} & {\bestval{6283.7 (61.26)}} & {\diffcolor{117.6}} \\ 
\cmidrule(lr){2-6}
                        & \multirow{2}{*}{Scenario 8} & Shifters & {\bestval{87.7 (2.54)}} & {\bestval{89.3 (2.20)}} & {\diffcolor{-1.6}} \\ 
                        &                             & Op. Time & {\bestval{3519.9 (47.36)}} & {\bestval{3431.0 (49.52)}} & {\diffcolor{88.9}} \\ 
\midrule
\multirow{6}{*}{PPO}    & \multirow{2}{*}{Scenario 6} & Shifters & {4.9 (1.84)} & {5.0 (1.85)} & {-0.1} \\
                        &                             & Op. Time & {1003.8 (75.47)} & {\bestval{949.9 (85.14)}} & {\diffcolor{54.0}} \\ 
\cmidrule(lr){2-6}
                        & \multirow{2}{*}{Scenario 7} & Shifters & {151.0 (3.07)} & {150.7 (2.80)} & {0.3} \\
                        &                             & Op. Time & {6405.7 (88.58)} & {6380.6 (83.50)} & {25.2} \\ 
\cmidrule(lr){2-6}
                        & \multirow{2}{*}{Scenario 8} & Shifters & {95.5 (3.73)} & {97.9 (3.52)} & {\diffcolor{-2.4}} \\
                        &                             & Op. Time & {3616.5 (80.31)} & {3532.6 (70.03)} & {\diffcolor{83.9}} \\
\midrule
\multirow{6}{*}{A2C}    & \multirow{2}{*}{Scenario 6} & Shifters & {\worstval{6.4 (3.85)}} & \worstval{5.3 (2.50)} & {1.0} \\ 
                        &                             & Op. Time & {\worstval{1084.7 (143.37)}} & {\worstval{1049.0 (123.74)}} & {35.7} \\ 
\cmidrule(lr){2-6}
                        & \multirow{2}{*}{Scenario 7} & Shifters & {\worstval{157.4 (3.09)}} & {157.1 (3.22)} & {0.3} \\ 
                        &                             & Op. Time & {6518.2 (84.91)} & {6451.8 (86.30)} & {\diffcolor{66.4}} \\
\cmidrule(lr){2-6}
                        & \multirow{2}{*}{Scenario 8} & Shifters & \worstval{107.9 (6.03)} & {\worstval{117.3 (6.32)}} & {\diffcolor{-9.3}} \\ 
                        &                             & Op. Time & {3699.7 (123.27)} & {\worstval{3753.2 (162.01)}} & {-53.5} \\ 
\midrule
\multirow{6}{*}{DQN}    & \multirow{2}{*}{Scenario 6} & Shifters & {4.4 (1.47)} & {4.3 (1.60)} & {0.1} \\
                        &                             & Op. Time & {1000.6 (61.30)} & {982.6 (62.53)} & {18.0} \\
\cmidrule(lr){2-6}
                        & \multirow{2}{*}{Scenario 7} & Shifters & {156.7 (5.06)} & \worstval{157.6 (8.08)} & {-0.9} \\ 
                        &                             & Op. Time & {\worstval{6577.9 (153.82)}}& {\worstval{6592.8 (260.21)}} & {-14.9} \\ 
\cmidrule(lr){2-6}
                        & \multirow{2}{*}{Scenario 8} & Shifters & {104.4 (6.99)} & {100.9 (6.97)} & {3.5} \\ 
                        &                             & Op. Time & {\worstval{3702.6 (131.09)}} & {3603.4 (83.71)} & {\diffcolor{99.3}} \\ 
\midrule
\multirow{6}{*}{QR-DQN} & \multirow{2}{*}{Scenario 6} & Shifters & {\bestval{3.9 (1.24)}} & {\bestval{3.5 (1.14)}} & {0.4} \\ 
                        &                             & Op. Time & {\bestval{983.8 (60.57)}} & {1001.0 (74.36)} & {-17.2} \\ 
\cmidrule(lr){2-6}
                        & \multirow{2}{*}{Scenario 7} & Shifters & {150.4 (3.72)} & {152.3 (2.79)} & \diffcolor{-1.8} \\
                        &                             & Op. Time & {6412.4 (114.80)} & {6492.9 (155.58)} & {\diffcolor{-80.5}} \\
\cmidrule(lr){2-6}
                        & \multirow{2}{*}{Scenario 8} & Shifters & {94.8 (4.11)} & {97.9 (4.11)} & {\diffcolor{-3.1}} \\
                        &                             & Op. Time & {3607.2 (98.07)} & {3628.6 (126.93)} & {-21.3} \\
\bottomrule
\end{tabular}
\end{table}

Experiments in multi-crane scenarios (6, 7, and 8), as shown in Figure \ref{fig:spgemc_optime_curves}, revealed similar algorithm rankings, with TRPO remaining dominant in complex settings and A2C underperforming. A detailed comparison of the single-agent (SPGE-MC) and multi-agent (SPAEC) formulations is summarized in Table \ref{tab:final_kpi_comparison_colored_full}, with statistically significant differences (t-test, p<0.05) highlighted.

For the two different formulations, single-agent and multi-agent, their impact becomes significant only in more complex scenarios (Scenario 7 and 8) and with certain algorithms. This impact cannot be generalized as one formulation being universally better. A trend we can observe is that TRPO is the algorithm most likely to exhibit notable performance differences between the two formulations in terms of both reducing shifters and operation time under distinct scenario-algorithm combinations. Moreover, the single-agent formulation tends to be more advantageous for optimizing shifters in most cases.

As discussed in Section \ref{sec:spge-mc}, due to SPAEC’s agent selection mechanism, the single-agent setup in SPGE-MC has more flexibility in crane selection when multiple cranes are available at a given step, which might be helpful for agent to learn a globally optimal policy for minimizing total shifters. While minimizing shifters and minimizing total operation time can become conflicting objectives beyond a certain optimization threshold, which requiring trade-offs, it might be easier for agent to prioritize shifter reducing since shifter-based rewards are denser.

\section{Conclusion}
This paper investigates the application of deep reinforcement learning to the Container Stowage Planning Problem (CSPP), comparing five algorithms (DQN, QR-DQN, A2C, PPO, TRPO) across various problem scales and formulations. Our primary objectives were minimizing shifters and, in multi-crane settings, total operation time. Through a series of experiments in our custom-designed environments, we found that while all algorithms perform comparably on simple problems, their effectiveness diverges significantly as complexity increases. A2C and value-based methods like DQN and QR-DQN proved susceptible to performance degradation and suboptimal solutions in complex scenarios. In contrast, PPO and TRPO demonstrated superior performance, with TRPO consistently achieving the best performance in the most challenging settings.

Furthermore, when jointly optimizing stowage and crane scheduling, we observed that the performance difference due to problem formulation was not significant in simple scenarios but became more noticeable in complex settings, though not always in a consistent manner. But overall, the single-agent formulation tends to perform better in reducing shifters during algorithm training.

In the future, this research can be extended by designing richer scenarios, such as increasing the problem scales, and implementing more realistic vessel hull structures, adding features like hatches and specialized zones, and developing more sophisticated models for operation time to reflect more aspects in real-world operations.

\begin{credits}
\subsubsection{\ackname} This study was supported by Leiden University and loadmaster.ai. The computations for this work were carried out using the resources of the Academic Leiden Interdisciplinary Cluster Environment (ALICE) provided by Leiden University.

\end{credits}


%
%
%
\clearpage
\bibliographystyle{splncs04}
\bibliography{mybibliography}

\begin{thebibliography}{10}
\providecommand{\url}[1]{\texttt{#1}}
\providecommand{\urlprefix}{URL }
\providecommand{\doi}[1]{https://doi.org/#1}

\bibitem{AMBROSINO200481}
Ambrosino, D., Sciomachen, A., Tanfani, E.: Stowing a containership: the master bay plan problem. Transportation Research Part A: Policy and Practice  \textbf{38}(2),  81--99 (2004). \doi{https://doi.org/10.1016/j.tra.2003.09.002}, \url{https://www.sciencedirect.com/science/article/pii/S0965856403000892}

\bibitem{avriel1993exact}
Avriel, M., Penn, M.: Exact and approximate solutions of the container ship stowage problem. Computers \& industrial engineering  \textbf{25}(1-4),  271--274 (1993)

\bibitem{azevedo2018solving}
Azevedo, A.T., de~Salles~Neto, L.L., Chaves, A.A., Moretti, A.C.: Solving the 3d stowage planning problem integrated with the quay crane scheduling problem by representation by rules and genetic algorithm. Applied Soft Computing  \textbf{65},  495--516 (2018)

\bibitem{bellman1966dynamic}
Bellman, R.: Dynamic programming. science  \textbf{153}(3731),  34--37 (1966)

\bibitem{1606.01540}
Brockman, G., Cheung, V., Pettersson, L., Schneider, J., Schulman, J., Tang, J., Zaremba, W.: Openai gym (2016)

\bibitem{cho2024developing}
Cho, J., Ku, N.: Developing a container ship loading-planning program using reinforcement learning. Journal of Marine Science and Engineering  \textbf{12}(10), ~1832 (2024)

\bibitem{cruz2015lower}
Cruz-Reyes, L., Hernandez~Hernandez, P., Melin, P., Mar-Ortiz, J., Fraire~Huacuja, H.J., Puga~Soberanes, H.J., Gonzalez~Barbosa, J.J.: Lower and upper bounds for the master bay planning problem. International Journal of Combinatorial Optimization Problems and Informatics  \textbf{6}(1),  42--52 (2015)

\bibitem{dabney2018distributional}
Dabney, W., Rowland, M., Bellemare, M., Munos, R.: Distributional reinforcement learning with quantile regression. In: Proceedings of the AAAI conference on artificial intelligence. vol.~32 (2018)

\bibitem{HE2023101864}
He, J., Zhang, L., Deng, Y., Yu, H., Huang, M., Tan, C.: An allocation approach for external truck tasks appointment in automated container terminal. Advanced Engineering Informatics  \textbf{55},  101864 (2023). \doi{https://doi.org/10.1016/j.aei.2022.101864}, \url{https://www.sciencedirect.com/science/article/pii/S1474034622003226}

\bibitem{henderson2018deep}
Henderson, P., Islam, R., Bachman, P., Pineau, J., Precup, D., Meger, D.: Deep reinforcement learning that matters. In: Proceedings of the AAAI conference on artificial intelligence. vol.~32 (2018)

\bibitem{hsu2021joint}
Hsu, H.P., Wang, C.N., Fu, H.P., Dang, T.T.: Joint scheduling of yard crane, yard truck, and quay crane for container terminal considering vessel stowage plan: An integrated simulation-based optimization approach. Mathematics  \textbf{9}(18), ~2236 (2021)

\bibitem{huang2020closer}
Huang, S., Onta{\~n}{\'o}n, S.: A closer look at invalid action masking in policy gradient algorithms. arXiv preprint arXiv:2006.14171  (2020)

\bibitem{jayawardana2022impact}
Jayawardana, V., Tang, C., Li, S., Suo, D., Wu, C.: The impact of task underspecification in evaluating deep reinforcement learning. Advances in Neural Information Processing Systems  \textbf{35},  23881--23893 (2022)

\bibitem{jensen2018container}
Jensen, R.M., Pacino, D., Ajspur, M.L., Vesterdal, C.: Container vessel stowage planning. Weilbach (2018)

\bibitem{jiang2021new}
Jiang, T., Zeng, B., Wang, Y., Yan, W.: A new heuristic reinforcement learning for container relocation problem. In: Journal of Physics: Conference Series. vol.~1873, p. 012050. IOP Publishing (2021)

\bibitem{kizilay2021comprehensive}
Kizilay, D., Eliiyi, D.T.: A comprehensive review of quay crane scheduling, yard operations and integrations thereof in container terminals. Flexible Services and Manufacturing Journal  \textbf{33}(1),  1--42 (2021)

\bibitem{mnih2016asynchronous}
Mnih, V., Badia, A.P., Mirza, M., Graves, A., Lillicrap, T., Harley, T., Silver, D., Kavukcuoglu, K.: Asynchronous methods for deep reinforcement learning. In: International conference on machine learning. pp. 1928--1937. PmLR (2016)

\bibitem{mnih2013playing}
Mnih, V., Kavukcuoglu, K., Silver, D., Graves, A., Antonoglou, I., Wierstra, D., Riedmiller, M.: Playing atari with deep reinforcement learning. arXiv preprint arXiv:1312.5602  (2013)

\bibitem{padakandla2021survey}
Padakandla, S.: A survey of reinforcement learning algorithms for dynamically varying environments. ACM Computing Surveys (CSUR)  \textbf{54}(6),  1--25 (2021)

\bibitem{stable-baselines3}
Raffin, A., Hill, A., Gleave, A., Kanervisto, A., Ernestus, M., Dormann, N.: Stable-baselines3: Reliable reinforcement learning implementations. Journal of Machine Learning Research  \textbf{22}(268), ~1--8 (2021), \url{http://jmlr.org/papers/v22/20-1364.html}

\bibitem{Reda_2020}
Reda, D., Tao, T., van~de Panne, M.: Learning to locomote: Understanding how environment design matters for deep reinforcement learning. In: Motion, Interaction and Games. MIG ’20, ACM (Oct 2020). \doi{10.1145/3424636.3426907}, \url{http://dx.doi.org/10.1145/3424636.3426907}

\bibitem{schulman2015trust}
Schulman, J., Levine, S., Abbeel, P., Jordan, M., Moritz, P.: Trust region policy optimization. In: International conference on machine learning. pp. 1889--1897. PMLR (2015)

\bibitem{schulman2017proximal}
Schulman, J., Wolski, F., Dhariwal, P., Radford, A., Klimov, O.: Proximal policy optimization algorithms. arXiv preprint arXiv:1707.06347  (2017)

\bibitem{shen2017deep}
Shen, Y., Zhao, N., Xia, M., Du, X.: A deep q-learning network for ship stowage planning problem. Polish Maritime Research  \textbf{24}(s3),  102--109 (2017)

\bibitem{10.5555/3312046}
Sutton, R.S., Barto, A.G.: Reinforcement Learning: An Introduction. A Bradford Book, Cambridge, MA, USA (2018)

\bibitem{terry2021pettingzoo}
Terry, J., Black, B., Grammel, N., Jayakumar, M., Hari, A., Sullivan, R., Santos, L.S., Dieffendahl, C., Horsch, C., Perez-Vicente, R., et~al.: Pettingzoo: Gym for multi-agent reinforcement learning. Advances in Neural Information Processing Systems  \textbf{34},  15032--15043 (2021)

\bibitem{:/content/books/9789211065923}
Trade, U., Development: Review of Maritime Transport 2024. United Nations, 2024 edn. (2024), \url{https://www.un-ilibrary.org/content/books/9789211065923}

\bibitem{van2023towards}
van Twiller, J., Grbic, D., Jensen, R.M.: Towards a deep reinforcement learning model of master bay stowage planning. In: International Conference on Computational Logistics. pp. 105--121. Springer (2023)

\bibitem{VANTWILLER2024841}
{van Twiller}, J., Sivertsen, A., Pacino, D., Jensen, R.M.: Literature survey on the container stowage planning problem. European Journal of Operational Research  \textbf{317}(3),  841--857 (2024). \doi{https://doi.org/10.1016/j.ejor.2023.12.018}, \url{https://www.sciencedirect.com/science/article/pii/S0377221723009517}

\bibitem{wei2021optimization}
Wei, L., Wei, F., Schmitz, S., Kunal, K.: Optimization of container relocation problem via reinforcement learning. Logistics Journal: Proceedings  \textbf{2021}(17) (2021)

\bibitem{williams1992simple}
Williams, R.J.: Simple statistical gradient-following algorithms for connectionist reinforcement learning. Machine learning  \textbf{8},  229--256 (1992)

\bibitem{wilson2000container}
Wilson, I.D., Roach, P.A.: Container stowage planning: a methodology for generating computerised solutions. Journal of the Operational Research Society  \textbf{51}(11),  1248--1255 (2000)

\bibitem{xia2020loading}
Xia, M., Li, Y., Shen, Y., Zhao, N.: Loading sequencing problem in container terminal with deep q-learning. Journal of Coastal Research  \textbf{103}(SI),  817--821 (2020)

\bibitem{zhao2018container}
Zhao, N., Guo, Y., Xiang, T., Xia, M., Shen, Y., Mi, C.: Container ship stowage based on monte carlo tree search. Journal of Coastal Research (83),  540--547 (2018)

\bibitem{zhao2020sim}
Zhao, W., Queralta, J.P., Westerlund, T.: Sim-to-real transfer in deep reinforcement learning for robotics: a survey. In: 2020 IEEE symposium series on computational intelligence (SSCI). pp. 737--744. IEEE (2020)

\bibitem{zheng2010effective}
Zheng, K., Lu, Z., Sun, X.: An effective heuristic for the integrated scheduling problem of automated container handling system using twin 40'cranes. In: 2010 second international conference on computer modeling and simulation. vol.~1, pp. 406--410. IEEE (2010)

\bibitem{zhou2022emerging}
Zhou, C., Zhu, S., Bell, M.G., Lee, L.H., Chew, E.P.: Emerging technology and management research in the container terminals: Trends and the covid-19 pandemic impacts. Ocean \& Coastal Management  \textbf{230},  106318 (2022)

\bibitem{zhu2020integer}
Zhu, H., Ji, M., Guo, W.: Integer linear programming models for the containership stowage problem. Mathematical Problems in Engineering  \textbf{2020}(1),  4382745 (2020)

\end{thebibliography}
%




\end{document}